\documentclass[article,reqno,twocolumn,3p]{elsarticle}
\usepackage{etex}
\usepackage{enumerate}
\usepackage{algorithm,algorithmic}
\usepackage{epic,eepic,graphicx}
\usepackage{multicol}
\usepackage{subcaption}
\usepackage{epstopdf}
\usepackage{amsmath,amssymb}
\usepackage{pgfplots}
\usepackage{bbm}
\usepackage{booktabs}
\usepackage{tabularx}
\usepackage{color}
\usepackage{hyperref}
\usepackage{colortbl}
\usepackage{pgfplotstable}
\usepackage{tikz}
\usepackage{chemfig}
\usepackage{rotating}
\usepackage{verbatim}
\usetikzlibrary{pgfplots.groupplots}
\usetikzlibrary{matrix,positioning}
\usetikzlibrary{external}
\tikzsetexternalprefix{Tikz/}
\pgfplotscreateplotcyclelist{colors}{%
	red,blue,teal,orange,cyan,black,violet,green!70!black,magenta,gray,yellow,brown}
\pgfplotsset{
	log x ticks with fixed point/.style={
		xticklabel={
			\pgfkeys{/pgf/fpu=true}
			\pgfmathparse{exp(\tick)}%
			\pgfmathprintnumber[fixed relative, precision=3]{\pgfmathresult}
			\pgfkeys{/pgf/fpu=false}
		}
	},
	log y ticks with fixed point/.style={
		yticklabel={
			\pgfkeys{/pgf/fpu=true}
			\pgfmathparse{exp(\tick)}%
			\pgfmathprintnumber[fixed,use comma,
			precision=3]{\pgfmathresult}
			\pgfkeys{/pgf/fpu=false}
		}
	}
}

\newlength\Colsep
\newcommand{\algname}{FSDA}
\newcommand{\osr}{CSR-SDA}
\newcommand{\ndsr}{SA-SDA}
\newcommand{\sr}{SR-SDA}

\begin{document}
	
\begin{frontmatter}

%\title{Fast semi-supervised discriminant analysis for binary classification of large data-sets\tnoteref{t1,t2}}
\title{Fast semi-supervised discriminant analysis for binary classification of large data-sets}
%\tnotetext[t1]{This document is a collaborative effort.}
%\tnotetext[t2]{The second title footnote which is a longer
%	longer then the first one and with an intention to fill
%	in up more then one line while formatting.}

%\author[deptcw,esat]{Joris Tavernier\corref{cor1}\fnref{fn1}}
\author[deptcw,esat]{Joris Tavernier\corref{cor1}}
\ead{Joris.Tavernier@cs.kuleuven.be}

\author[esat]{Jaak Simm}
\ead{jaak.simm@esat.kuleuven.be}

\author[deptcw]{Karl Meerbergen}
\ead{Karl.Meerbergen@cs.kuleuven.be}

\author[japh]{Joerg Kurt Wegner}
\ead{jwegner@its.jnj.com}

\author[japh]{Hugo Ceulemans}
\ead{hceulema@its.jnj.com}

\author[esat]{Yves Moreau}
\ead{moreau@esat.kuleuven.be}

\cortext[cor1]{Corresponding author}

%\fntext[fn1]{IWT: IWT130406. ExaScience.Life HPC, FWO: G079016N}
%\fntext[fn2]{Another author
%	footnote, but a little more
%	longer.}
%\fntext[fn3]{Yet	another	author	footnote.Indeed,you can have
%	any number of author footnotes.}

\address[deptcw]{Department of Computer Science, KU Leuven, Celestijnenlaan 200A, B-3001 Heverlee - Belgium}
\address[esat]{ESAT-STADIUS, KU Leuven, Kasteelpark Arenberg 10, bus 2446, B-3001 Heverlee - Belgium}
\address[japh]{Janssen Pharmaceutica, Turnhoutseweg 30,
	B-2340 Beerse - Belgium}

\begin{abstract}
	High-dimensional data requires scalable algorithms. We propose and analyze three scalable and related algorithms for semi-supervised discriminant analysis (SDA). These methods are based on Krylov subspace methods which exploit the data sparsity and the shift-invariance of Krylov subspaces. In addition, the problem definition was improved by adding centralization to the semi-supervised setting. The proposed methods are evaluated on a industry-scale data set from a pharmaceutical company to predict compound activity on target proteins. The results show that SDA achieves good predictive performance and our methods only require a few seconds, significantly improving computation time on previous state of the art.  
\end{abstract}

\end{frontmatter}
%\maketitle

\section{Introduction}
In fields such as data mining, information retrieval or pattern recognition, dimensionality reduction of high-dimensional data is often used for classification. This assumes that data can be accurately represented in a lower dimensional manifold. For instance, principal component analysis (PCA) \cite{hotelling1933analysis,bishop2006pattern} is a widely used unsupervised feature extraction method. In addition, linear discriminant analysis (LDA) \cite{bishop2006pattern,mclachlan2004discriminant,friedman2001elements} is a supervised method using dimensionality reduction to classify data.

For certain data sets, only some sample labels are known. For certain applications, verifying data point labels is extremely expensive. These applications can benefit from models that incorporate both labeled and unlabeled data in a semi-supervised setting \cite{zhu2009introduction}. Semi-supervised learning is built on the assumption that taking advantage of the inherent structure of unlabeled and labeled data aids in a classification task.

The difficulty and cost of labeling samples for certain applications stimulated surging interest in semi-supervised learning. This interest has led to several different methods of incorporating information of the unlabeled samples to aid the supervised classification task such as Semi-supervised Orthogonal Discriminant Analysis via label propagation (SODA) \cite{nie2009semi}, SEmi-supervised Local Fisher discriminant analysis (SELF) \cite{sugiyama2010semi}, Trace Ratio based Flexible Semi-supervised Discriminant Analysis, \cite{wang2007trace,huang2012semi}, Flexible Manifold Embedding (FME)\cite{nie2010flexible} and Semi-supervised Linear Discriminant Analysis \cite{wang2016semi}. Most of these methods try to incorporate geometric information of the unlabeled data and potentially estimate the labels to incorporate the unlabeled data in a known supervised method. 

Target prediction in chemogenomics is an example for which the semi-supervised setting could prove beneficial. Chemical compounds are modeled by combinations of molecules. Depending on the specific target protein, the compound can be active or inactive\footnote{In general, the chemical compound or one of the molecules the compound exists of can be active on the protein or not.}. However, due to existing drugs and research, only a few activity pairs are known. Labeling these activity pairs in vitro is expensive. Predicting which molecule affects a protein aids in the development of new drugs \cite{keiser2007relating,keiser2009predicting}. This is a binary classification task. A possible feature set of a compound are the binary substructure fingerprints, which are the substructures occurring in a compound \cite{rogers2010extended}. The resulting data matrix is a high-dimensional sparse binary matrix. 

Scalable algorithms and high performance computing are required for such high-dimensional data. Algorithms based on QR-decomposition or singular value decomposition have cubic computational complexity and are unfeasible. This eliminates several state-of-the-art semi-supervised learning methods mentioned before due to computational workload. LDA has been studied for large-scale data and scalable methods are available in the literature \cite{cai2008srda,zhu2014rayleigh}. More specifically, Spectral regression discriminant analysis (SRDA) \cite{cai2008srda} solves LDA efficiently and is a special case of spectral regression (SR) \cite{cai2007spectral} consisting of two phases: spectral analysis (SA) of an embedded graph and regression. Using manifold regularization \cite{belkin2006manifold}, LDA was extended to the semi-supervised setting leading to semi-supervised discriminant analysis (SDA) \cite{cai2007semi}. 

Here, we present data centralization to improve the problem definition of SDA and how to efficiently implement the centralization. Then, we use the shift-invariance property of Krylov subspace methods to find solutions for different regularization parameters in a computationally feasible manner. We suggest three scalable algorithms for SDA and compare predictive performance and execution time. These algorithms are applied on real-life large-scale data from the database of Janssen Pharmaceutica. Our main contributions are%: an efficient implementation of SDA using the shift invariance property of Krylov subspace methods for multiple regularization parameters; improving the SDA problem definition by adding centralization; applying spectral regression with centralization to SDA; showing how to classify data using spectral analysis as a standalone method in the semi-supervised setting; comparison of different algorithms to solve SDA; applying semi-supervised learning in the form of SDA to target prediction in chemogenomics. 
%\begin{itemize}
%	\item an efficient implementation of SDA using the shift invariance property of Krylov subspace methods for multiple regularization parameters;
%	\item improving the SDA problem definition by adding centralization;
%	\item applying spectral regression with centralization to SDA;
%	\item showing how to classify data using spectral analysis as a standalone method in the semi-supervised setting;
%	\item comparison of different algorithms to solve SDA;
%	\item applying semi-supervised learning in the form of SDA to target prediction in chemogenomics. 
%\end{itemize}
\begin{itemize}
	\item an efficient implementation of SDA using the shift invariance property of Krylov subspace methods for multiple regularization parameters and additionally improve the SDA problem definition by adding centralization;
	\item applying spectral regression with centralization to SDA and proposing classification of data using spectral analysis as a standalone method in the semi-supervised setting;
	\item applying semi-supervised learning in the form of SDA to target prediction in chemogenomics and comparing different algorithms for solving SDA.
\end{itemize}

Section \ref{sec:LDA} presents a brief review of LDA and its properties and Section \ref{sec:SDA} builds on the properties of LDA to introduce SDA. In Section \ref{sec:LA} the necessary tools are explained to apply SDA on large data sets and our scalable algorithms are defined in Section \ref{sec:proposed}. Finally SDA is applied on two data sets from chemogenomics in Section \ref{sec:results}.     
\section{Discriminant analysis}

In this section, we review LDA and present an existing scalable algorithm for LDA. Next, we detail a semi-supervised extension for LDA, named SDA. Finally, centralization is detailed for LDA and provided for SDA. 
\begin{table}[ht]
	\renewcommand{\arraystretch}{0.8}
	\begin{center}
		\scalebox{1.0}{
			\begin{tabular}{c | p{6cm}}\toprule
				Symbol& Description \\ \midrule
				 $D$& the number of features \\ 
				 $N$& the total number of data samples\\
				 $l$& the number of labeled data samples\\				 
				 $C$& the number of classes\\
				 $N_c$& the number of data samples in class $c$\\
				 $\textbf{x}_i$& $i$-th data sample \\
				 $X$& the data matrix $X\in\mathbb{R}^{N\times D}$\\
				 $S_B$& the between-class scatter matrix\\
				 $S_W$& the within-class scatter matrix\\
				 $S_T$& the total scatter matrix\\
				 $L$& the laplacian matrix\\
				 $\textbf{z}$& eigenvector of the spectral analysis eigenvalue problem, $\textbf{z}\in\mathbb{R}^{N\times 1}$\\
				 $\textbf{w}$& eigenvector of the spectral analysis eigenvalue problem projected onto the range of $X$, $\textbf{w}\in\mathbb{R}^{D\times 1}$\\
				 SR& spectral regression \\
				 SA& spectral analysis\\
				 CSR& centralized spectral regression\\
				 SDA& semi-supervised discriminant analysis\\
				 \bottomrule
		\end{tabular}}
	\end{center}
	\caption{Notations and descriptions used in the paper.}\label{tab:notation}
\end{table} 

\subsection{Linear discriminant analysis}\label{sec:LDA}
LDA \cite{bishop2006pattern,mclachlan2004discriminant,friedman2001elements,cai2008srda,cai2007semi} is a supervised dimensionality reducing method and finds the directions that maximally separate the different classes while minimizing the spread within one class. Suppose we have $C$ classes and $N$ samples $\textbf{x}_i$ of dimension $D$. The data matrix $X\in\mathbb{R}^{N\times D}$ is defined as 
%\begin{displaymath}X=\begin{pmatrix}
 %\textendash\textbf{x}_1\textendash\\\vdots\\\textendash\textbf{x}_i\textendash \\ %\vdots \\\textendash\textbf{x}_N\textendash \end{pmatrix} \end{displaymath}
 \begin{displaymath}X=[
 \textbf{x}_1,\ldots,\textbf{x}_i,  \ldots ,\textbf{x}_N ]^T \end{displaymath}
  where each row corresponds to a sample. We define $S_B$ as the between-class-scatter matrix \begin{displaymath}
  S_B=\sum_{c=1}^C N_c (\boldsymbol{\mu}_c-\boldsymbol{\mu})^T(\boldsymbol{\mu}_c-\boldsymbol{\mu})
  \end{displaymath} with $N_c$ the number of data points in class $c$, $\boldsymbol{\mu}$ the mean sample and $\boldsymbol{\mu}_c$ the mean sample of class $c$. The within-class-scatter matrix $S_W$ is defined by \begin{displaymath}
  S_W=\sum_{c=1}^C \sum_{n=1}^{N_c} (\textbf{x}_n-\boldsymbol{\mu}_c)^T(\textbf{x}_n-\boldsymbol{\mu}_c).
  \end{displaymath} The LDA objective function is then \begin{displaymath}\max_{\textbf{w}\in\mathbb{R}^{N}} \frac{\textbf{w}^TS_B\textbf{w}}{\textbf{w}^TS_W\textbf{w}}.  \end{displaymath} Using the total scatter matrix $S_T$ \begin{align*}
  S_T&=\sum_{n=1}^{N} (\textbf{x}_n-\boldsymbol{\mu})^T(\textbf{x}_n-\boldsymbol{\mu})\\
  &=S_B+S_W
  \end{align*} the objective function can then equivalently be expressed as  \begin{displaymath}\max_{\textbf{w}\in\mathbb{R}^{N}} \frac{\textbf{w}^TS_B\textbf{w}}{\textbf{w}^TS_T\textbf{w}} . \end{displaymath}Since $S_B$ and $S_T$ are positive semi-definite, the optimization problem can be reformulated as finding the large eigenvalues and eigenvectors of \begin{equation}S_B\textbf{w}_i=\lambda_i(S_T+\beta I)\textbf{w}_i \label{eqn:dataSwEig} \end{equation} which we call the LDA eigenvalue problem. The regularization parameter $\beta$ is introduced due to possibility of singularity of the total scatter matrix and as a countermeasure against overfitting. Suppose the data matrix $X$ is centered and sorted according to the class of the data points. $$X=[X_1;X_2;\dots;X_C]$$ where $X_i$ are the data samples of class $i$. The between-class covariance matrix can then be simplified to \begin{displaymath}
S_B=X^TWX
\end{displaymath} with $W$ an $N\times N$ block diagonal matrix with C matrices ($W_1,\dots,W_C$) on its diagonal and each value of matrix $W_c\in \mathbb{R}^{N_c\times N_c}$ is equal to $1/N_c$ and $N_c$ is the number of samples in class $c$. The total scatter matrix can be written as $S_T=X^TX$ and the LDA eigenvalue problem is then
  \begin{equation}X^TWX\textbf{w}_i=\lambda_i(X^TX+\beta I)\textbf{w}_i \label{eqn:dataEig}. \end{equation}
\subsection{Spectral regression}
LDA has been investigated for large-scale data. Here we present one state-of-the-art algorithm for LDA for large scale data named Spectral regression discriminant analysis (SRDA) \cite{cai2008srda}. SRDA can actually be interpreted as a part of a graph embedding analysis technique called spectral regression \cite{cai2007spectral}.  
\subsubsection{Spectral regression discriminant analysis}
The LDA eigenvalue problem (\ref{eqn:dataEig}) can be interpreted as the projection of the eigenvalue problem \begin{equation}W\textbf{z}_i=\lambda_i\textbf{z}_i\label{eqn:wEig} \end{equation} on the range of $X$ \cite{cai2008srda}. This eigenvalue problem is quite simple due to the block diagonal structure of $W$ where values in one block are identical. There are two eigenvalues: $1$ with multiplicity $C$ and $0$ with multiplicity $N-C$\cite{cai2008srda}. In general, the space spanned by the eigenvectors with eigenvalue $1$ is the same as the space spanned by the vectors \begin{equation}
\textbf{z}_c=(\underbrace{0,\ldots,0}_{\sum_{m=1}^{c-1}N_m},\underbrace{1,\ldots,1}_{N_c},\underbrace{0,\ldots,0}_{\sum_{m=c+1}^{C}N_m})^T \label{eqn:Weigz}
\end{equation} due to the special block diagonal structure of $W$.

In order to gain some insight in eigenvalue problem (\ref{eqn:wEig}), we will consider the binary case. If we only have two classes, then the matrix $W$ has rank $2$. Taking a linear combination of $\textbf{z}_1$ and $\textbf{z}_2$ in (\ref{eqn:Weigz}), two eigenvectors with eigenvalue $1$ of eigenvalue problem (\ref{eqn:wEig}) are
\begin{equation*}
\textbf{z}_{nd}=\begin{pmatrix}
1\\\vdots\\1\\1\\\vdots\\1 \end{pmatrix}	,\ \textbf{z}_d=\begin{pmatrix}
1\\\vdots\\1\\-1\\\vdots\\-1 \end{pmatrix} \label{eqn:ndev}
\end{equation*}
where $\textbf{z}_d$ has value $1$ for the samples of class $1$ and value $-1$ for the samples of class $2$. It is obvious that eigenvector $\textbf{z}_{nd}$ with all values equal to $1$ has no discriminative information. The eigenvector $\textbf{w}_d$, with $X\textbf{w}_d=\textbf{z}_d$, is the direction that maximizes the LDA objective function when $\textbf{z}_d\in \text{Range}(X)$. In general, the eigenvalue associated with $\textbf{w}_d$ is smaller but close to $1$ since it is the projection onto the range of $X$ of the eigenvector $\textbf{z}_d$ with eigenvalue $1$. 

SRDA uses the knowledge that the space spanned by the eigenvectors $\textbf{z}_c$ (\ref{eqn:wEig}) includes the non-discriminative eigenvector $\textbf{z}_{nd}$. SRDA orthogonalizes the vectors $\textbf{z}_c$ with respect to $\textbf{z}_{nd}$, resulting in $C-1$ discriminative eigenvectors $\textbf{z}_1,\dots,\textbf{z}_{C-1}$ of eigenvalue problem (\ref{eqn:wEig}). SRDA then solves the linear systems $X\textbf{w}_c=\textbf{z}_c$ for $c=1,\dots,C-1$ in a regularized least-squares (RLS) manner, resulting in the $C-1$ eigenvectors of the LDA eigenvalue problem which are the directions that maximize the LDA objective function.  

\subsubsection{Spectral regression}
SRDA fits in the spectral regression (SR) framework \cite{cai2007spectral}. SR provides a graph embedding interpretation to different dimensionality reduction techniques such as Semantic Subspace Projection \cite{yu2006learning} or Locality Preserving Projections \cite{he2004locality}. Based on a graph, a symmetric matrix $W$ represents the weights of the edges between two nodes or in our case data samples. Note that for LDA the resulting graph consists of the data points and two points are connected by an edge with weight $1/N_c$ if they have the same class $c$.

Spectral analysis (SA) of the graph computes the eigenvectors associated with the largest eigenvalues of \begin{equation}
W\textbf{z}=\lambda D\textbf{z} \label{eqn:saeig}
\end{equation}
with D a diagonal matrix and $D_{ii}=\sum_{n=1}^N W_{in}$. As shown by Cai et al \cite{cai2007spectral}, this corresponds to penalizing two connected nodes $i,j$ with a large difference $|z_i-z_j|$. For new data samples a linear function is chosen such that $z_i=\textbf{x}_i\textbf{w}$, resulting in $\textbf{z}=X\textbf{w}$. Eigenvalue problem (\ref{eqn:saeig}) then results in 
\begin{equation}
X^TWX\textbf{w}=\lambda X^TDX\textbf{w} \label{eqn:saxeig}
\end{equation} 
which for LDA is similar to (\ref{eqn:dataEig}) without regularization.

SR thus first computes the eigenvectors in (\ref{eqn:saeig}) using the Lanczos method \cite{demmel1997applied} and is called the spectral analysis phase. Note that for SRDA these eigenvectors were computed analytically and no eigenvalue solve is required. SR secondly solves the linear system $\textbf{z}=X\textbf{w}$ in a regularized least-squares manner $$\textbf{w}=(X^TX+\beta I)^{-1}X^T\textbf{z}$$to find the desired directions $\textbf{w}$ in (\ref{eqn:dataEig}) or (\ref{eqn:saxeig}) during the regression phase. 
   
\subsection{Semi-supervised discriminant analysis}\label{sec:SDA}
In the semi-supervised setting, the class of each data point is not always known and we will refer to the data points where the class is known as labeled data and unlabeled data otherwise. For both the unlabeled and labeled data, their structure in the feature space, $\textbf{x}_i$, is known. Semi-supervised learning combines both the class information of the labeled data and the structure in the feature space of labeled and unlabeled data. SDA \cite{cai2007semi} uses manifold regularization to extend LDA to the semi-supervised setting. Information of the geometric proximity of the data points in the feature space is contained in the regularizer.

Using a binary similarity matrix $S\in\mathbb{R}^{N\times N}$, where $s_{ij}=1=s_{ji}$ if two samples are similar and $0$ otherwise, a natural regularizer $R(\textbf{w})$ for LDA can be defined as \begin{align*}
R(\textbf{w})&=\sum_{i=1}^{n}\sum_{j=1}^{n}s_{ij}(\textbf{w}^T\textbf{x}_i-\textbf{w}^T \textbf{x}_j)^2 \label{eqn:sdaregul}\\
&=2\sum_{i=1}^{n}\left((\textbf{w}^T\textbf{x}_i)^2d_i-\sum_{j=1}^{n}s_{ij}\textbf{w}^T\textbf{x}_i\textbf{w}^T\textbf{x}_j\right)\\
&=2\textbf{w}^TX^T(D-S)X\textbf{w}
\end{align*}
where $d_i$ is the sum of the $i$th row of $S$ and $D\in \mathbb{R}^{n\times n}$ the diagonal matrix with $d_i$ on its diagonal. The matrix $L=D-S$ is the Laplacian matrix of $S$. Suppose the number of labeled data is $\ell$, the SDA eigenvalue problem is then 
\begin{equation}
\begin{split}
\underbrace{X^T\begin{pmatrix}
	W & 0\\0&0
	\end{pmatrix}X}_{\text{\small Between-labeled-class-scatter}}\textbf{w}_i=\\
\lambda_i\ [\underbrace{ X^T(1-\alpha)\begin{pmatrix}
	I_\ell&0\\0&0\end{pmatrix} X}_{\text{\small Total-labeled-scatter}}+\\
\underbrace{ X^T\alpha L X}_{\text{\small Manifold regularization}}+\\
\underbrace{\beta I}_{\text{\small Regularization}}]\ \textbf{w}_i \label{eqn:SDAeig}
\end{split}
\end{equation}
where the first $\ell$ rows of X are the labeled data points and $W\in\mathbb{R}^{\ell\times \ell}$ the same matrix as for LDA. The regularization parameter $\alpha$ is limited to the interval $[0,1]$ and resembles how much weight is given to the unsupervised information. Note that the information of the unlabeled data is only used in the Laplacian matrix $L$. SDA can be interpreted under the spectral regression framework and its SA formulation (\ref{eqn:saeig}) is
\begin{equation}
\begin{pmatrix}
W & 0\\0&0
\end{pmatrix}\textbf{z}=\lambda\left[(1-\alpha)\begin{pmatrix}
I_\ell&0\\0&0\end{pmatrix} + \alpha L\right]\textbf{z} \label{eqn:sasdaeig} 
\end{equation}
and is followed by the usual regression phase. 
\subsection{Similarity matrix}
Semi-supervised learning is based on the assumption that similar data points in the feature space are more likely to have the same class. Defining the similarity between data points is an important part of the semi-supervised setting. SDA uses a binary similarity matrix but this is not a necessary condition. For a binary similarity matrix, the first approach we considered is creating a $k$-nearest neighbor graph \cite{cai2007semi,yu2006learning}. 

The value of $s_{ij}\in S$ is $1$ if there exists an undirected edge in the $k$-nearest neighbor graph from data point $i$ to data point $j$. The $k$-nearest neighbor graph is built using a distance measure in the feature space. The distance for each data point is calculated to all the other data points. The data point is then connected by an undirected edge to all the other points maximally separated by the distance of the $k$-nearest neighbor.  

The second approach is to create a binary similarity matrix by thresholding. Using a distance measure a threshold is defined for all data points and if the distance from data point $i$ to data point $j$ is smaller than the threshold, the data points are similar and $s_{ij}=s_{ji}=1$. Note that using a threshold instead of the $k$-nearest neighbors graph may lead to samples having no neighbors at all.  

\subsection{Centralization}\label{sec:centralization}
Centralization, as for PCA, plays an important role in LDA. We will first look into the details of the influence of centralization for LDA and extend the same idea to SDA.  
\subsubsection{Centralization for LDA}
The non-discriminative eigenvector $\textbf{z}_{nd}$ can be removed from the LDA eigenvalue problem (\ref{eqn:dataEig}) by centralization of the data \cite{cai2008srda,zhu2014rayleigh}. If $\textbf{z}_{nd} \in \text{Range}(X)$, then the corresponding vector $\textbf{w}_{nd}$, with $X\textbf{w}_{nd}=\textbf{z}_{nd}$, is an eigenvector of the LDA eigenvalue problem $$X^TWX\textbf{w}_{nd}=\lambda_i (X^TX\textbf{w}_{nd}+\beta\textbf{w}_{nd}).$$
Note that due to the structure of $W$, \begin{eqnarray*}
	X^TWX\textbf{w}_{nd}&=X^TW\textbf{z}_{nd}\\
	&=X^T\textbf{z}_{nd}\\
	&=N\boldsymbol{\mu}
\end{eqnarray*} with $\boldsymbol{\mu}=X^T\textbf{z}_{nd}/N$ the mean sample in the feature space. If the data is centralized, the non-discriminative eigenvector $\textbf{w}_{nd}$ is moved to the common null space of $S_B$ and $S_T$ of the LDA eigenvalue problem (\ref{eqn:dataSwEig}). 
\subsubsection{Centralization for SDA}
The centralization trick from LDA to remove the non-discriminative eigenvector can be applied for SDA as well. Consider the vector $\textbf{z}_{nd}\in \mathbb{R}^{N\times 1}$ with all values equal to $1$. From the definition of the Laplacian matrix $L$, we know that $L\textbf{z}_{nd}=\textbf{0}$. Therefore, $\textbf{z}_{nd}$ is an eigenvector of the SA eigenvalue problem for SDA
\begin{equation}
\begin{pmatrix}
W & 0\\0&0
\end{pmatrix}\textbf{z}_{nd}=\lambda_{nd}\left[(1-\alpha)\begin{pmatrix}
I_\ell&0\\0&0\end{pmatrix}\textbf{z}_{nd} + \alpha L\textbf{z}_{nd}\right] 
\end{equation}
with $\lambda_{nd}=1/(1-\alpha)$. The SDA eigenvalue problem can be interpreted as a projection of this eigenvalue problem onto the range of $X$. As in LDA, we have
\begin{displaymath}
X^T\begin{pmatrix}
W & 0\\0&0
\end{pmatrix}\textbf{z}_{nd}=\ell\boldsymbol{\mu}_\ell 
\end{displaymath}  
with $\boldsymbol{\mu}_\ell$ the mean of the labeled data in the feature space. To remove the non-discriminative eigenvector, we centralize the data around the mean of the labeled data since the Laplacian does not influence the eigenvector $\textbf{w}_{nd}$, found by solving the system $X\textbf{w}_{nd}=\textbf{z}_{nd}$. This is an improvement to the original formulation of SDA \cite{cai2007semi}. 

\section{Large-scale data}\label{sec:LA}
Exact solutions for high dimensional data are costly to calculate and are no longer feasible. First we look at an iterative method to solve a large generalized singular eigenvalue problem in Section \ref{sec:eig} and next we look at Krylov subspace methods to solve linear systems in Section \ref{sec:Krylov}. 

\subsection{Subspace iteration}\label{sec:eig}
For the SDA eigenvalue problem, we have to solve a singular generalized eigenvalue problem \begin{equation}
A\textbf{v}=\lambda B\textbf{v} \label{eqn:geneig}
\end{equation} where $A$ is of low rank ($C-1$). It is known that subspace iteration has geometric asymptotic convergence for eigenvalue $\lambda_i$ with ratio $\lambda_{j+1}/\lambda_i$ for $i=1,\dots, j$ with $j$ the dimension of the subspace \cite{parlett1998symmetric}. Choosing $j=C-1$ leads to instant convergence due to the low rank of $A$ ($\lambda_{j+1}=0$). The subspace iteration algorithm applied to eigenvalue problem (\ref{eqn:geneig}) is given in Algorithm \ref{alg:subspace} with R randomly generated from a uniform distribution between $-1$ and $1$. 
\begin{algorithm}
	\caption{Subspace iteration for (\ref{eqn:geneig})}
	\label{alg:subspace}
	\begin{algorithmic}[1]
		\STATE Initialize $R\in \mathbb{R}^{D\times C-1}$ randomly
		\STATE Solve $V$ from $BV=AR$
		\RETURN V
	\end{algorithmic}
\end{algorithm} 

The computational bottleneck is solving the linear system with $B$. When $B$ is high-dimensional this can no longer be done exactly in a feasible manner and iterative methods are used. The convergence of subspace iteration is no longer guaranteed in one iteration. In general, many applications such as SDA require only an approximate solution and for these applications it is not necessary to solve the problem to machine precision. 
 
\subsection{Linear systems}\label{sec:Krylov}
We use Krylov subspace based methods for solving a linear system $B\textbf{w}=\textbf{b}$ on line 2 of Algorithm \ref{alg:subspace}. A Krylov base $\mathcal{K}$ is an orthogonal basis determined by the space of a sequence of matrix-vector multiplications $\mathcal{K}=\text{span}\{\textbf{b},B\textbf{b},\dots, B^{k-1} \textbf{b}\}$. A nice property of these methods is that if the matrix $B$ is sparse the cost of a matrix-vector multiplication is linear with the number of nonzero entries in the matrix $B$. For the SDA eigenvalue problem (\ref{eqn:SDAeig}), the matrix $B$ is a product of sparse matrices, which we do not compute explicitly since this generally leads to a dense matrix. This subsection is technical and more detailed information can be found in the literature \cite{demmel1997applied,trefethen1997numerical}.

\subsubsection{Solving the symmetric linear system}
For symmetric matrices (which is the case in SDA) the idea is to find a Krylov base $\mathcal{K}$ such that \begin{equation}
\mathcal{K}^*B\mathcal{K}=T \label{eqn:kryT}
\end{equation} with $$T=\begin{bmatrix}
\alpha_1& \gamma_1& & \\
\gamma_1&\ddots& \ddots & \\
&\ddots& \ddots &\gamma_{k-1} \\
& &\gamma_{k-1}&\alpha_k
\end{bmatrix}$$ a tridiagonal matrix. 

The approximate solution of the linear system $B\textbf{w}=\textbf{b}$ is found by searching the space spanned by the vectors of the Krylov space $\mathcal{K}$. One way to find an approximate solution is to make the residual orthogonal to the Krylov base ($(B+\beta I)\textbf{w}_k-\textbf{b} \perp \mathcal{K} $). This is true when $\textbf{w}_k=\mathcal{K}\textbf{z}_k$ where $\textbf{z}_k$ is the solution of the linear system\begin{equation}
(T+\beta I)\textbf{z}_k=||\textbf{b}||_2\textbf{e}_1\label{eqn:lanczos}\end{equation}  of dimension $k$. If the matrix is positive-definite, the solution minimizes the error under the $B$-norm. This method is the Lanczos algorithm to solve a linear system and is related to the conjugate gradient method (CG, Algorithm \ref{alg:cg}) for which they use a Cholesky factorization of $T$ to define a recursive process \cite{demmel1997applied}.
\subsubsection{Regularization}
Singular eigenvalue problems are well understood for small scale matrices \cite{demmel1993generalized,demmel1993generalized2}. However, these algorithms are unfeasible for large scale problems, since singular subspace deflation, removing the null space from both matrices, is required and has cubic complexity. In contrast, adding a regularization parameter to matrix $B$ (\ref{eqn:geneig}) is computationally cost-effective, equally well-defined and offers a countermeasure against overfitting.

Replacing $B$ by $B+\beta I$ makes the eigenvalue problem well defined for which we can use a standard method, such as subspace iteration. If $\beta$ is small, the eigenvalues and eigenvectors of interest do not change significantly \cite{tikhonov1977methods}. In addition, the singularity of $B$ is inexact, namely, the singular values quickly decay, but they do not reach zero. These singular values are the result of `noise' in the data or a mismatch in the imposed classification. The effect of this `noise' is reduced by adding the regularization parameter $\beta$.

In general, the best solution is chosen from a range of $\beta$ values in the terms of a performance measure. Krylov bases are invariant to shifts or regularization parameters and exploiting this property is very efficient \cite{van2004accurate,frommer1999fast,jegerlehner1996krylov}. In the previous discussion, the Krylov vectors are independent of the value of $\beta$ and the tridiagonal matrix $T$ (\ref{eqn:kryT}) is not influenced by $\beta$. This means that the computationally expensive part, that is, building the Krylov base, can be reused for several $\beta$ values instead of restarting the process for each $\beta$ value. The parameter $\beta$ only plays a role in solving the small linear system in (\ref{eqn:lanczos}). The conjugate gradient method for the shifted linear systems of the form $$(B+\beta I)\textbf{w}=\textbf{b}$$ is given in Algorithm \ref{alg:Shifted_cg} \cite{jegerlehner1996krylov,sciarracgalgorithm}. Superscript $\beta$ means that the variable is stored for all different regularization parameters $\beta$ in such a way that scalars ($\gamma$) become vectors ($\boldsymbol{\gamma}^\beta$) and vectors ($\textbf{w}$) change into matrices ($W^\beta$).  

\begin{figure*}
	\tikzsetnextfilename{algorithms}
\begin{minipage}[t]{.4\textwidth}
	\vspace{0pt} 
\centering 
\begin{algorithm}[H]
	\caption{Conjugate gradient}
	\label{alg:cg}
	\begin{algorithmic}[1]
		\STATE Initialize: \begin{itemize}
			\item $\textbf{r}_0=\textbf{p}_0=\textbf{b}$
			\item $\textbf{w}=\textbf{0}$ 
		\end{itemize} 
		\FOR{$i=0,\dots, k$}
		\STATE $\textbf{q}_i= B\textbf{p}_i$
		\STATE $\gamma_i=-\frac{\langle\textbf{r}_i,\textbf{r}_i\rangle}{\langle\textbf{p}_i,\textbf{q}_i\rangle}$
		\STATE $\textbf{w}_{i+1}=\textbf{w}_i-\gamma_i \textbf{p}_i$
		\STATE $\textbf{r}_{i+1}=\textbf{r}_i+\gamma_i\textbf{q}_i$
		\IF{$||\textbf{r}_{i+1}||<tol$} 
		\STATE Break
		\ENDIF 
		\STATE $\alpha_{i+1}=\frac{\langle\textbf{r}_{i+1},\textbf{r}_{i+1}\rangle}{\langle\textbf{r}_i,\textbf{r}_i\rangle}$
		\STATE $\textbf{p}_{i+1}=\textbf{r}_{i+1}+\alpha_{i+1}\textbf{p}_i$
		\ENDFOR
	\end{algorithmic}
\end{algorithm}
\end{minipage}%
\begin{minipage}[t]{.2\textwidth}
\end{minipage}
\begin{minipage}[t]{.4\textwidth}
\centering 	
	\vspace{0pt}

\begin{algorithm}[H]
	\caption{Shifted conjugate gradient}
	\label{alg:Shifted_cg}
	\begin{algorithmic}[1]
		\STATE Initialize: \begin{itemize}
			\item {\color{blue!60!black}$\alpha_0=0$},{\color{blue!60!black} $W_0^\beta=\textbf{0}$},  {\color{blue!60!black}$\boldsymbol{\alpha}^\beta_0=\textbf{0}$ }
			\item $\textbf{r}_0=\textbf{b}$, $\textbf{p}_0=\textbf{b}$,  {\color{blue!60!black}$P_0^\beta=\textbf{b}$} 
			\item {\color{blue!60!black}$\gamma_{-1}=1$}, {\color{blue!60!black}$\boldsymbol{\zeta}_{-1}^\beta=\textbf{1}$},  {\color{blue!60!black}$\boldsymbol{\zeta}_{0}^\beta=\textbf{1}$}  
		\end{itemize} 
		\FOR{$i=0,\dots, k$}
		\STATE $\textbf{q}_i= B\textbf{p}_i$
		\STATE $\gamma_i=-\frac{\langle\textbf{r}_i,\textbf{r}_i\rangle}{\langle\textbf{p}_i,\textbf{q}_i\rangle}$
		{\color{blue!60!black}\STATE $\boldsymbol{\zeta}_{i+1}^\beta =\frac{\boldsymbol{\zeta}_{i-1}^\beta\boldsymbol{\zeta}_{i}^\beta\gamma_{i-1} } {\gamma_i\alpha_i(\boldsymbol{\zeta}_{i-1}^\beta-\boldsymbol{\zeta}_{i}^\beta)+\boldsymbol{\zeta}_{i-1}^\beta\gamma_{i-1}(1-\boldsymbol{\beta}\gamma_i)} $
		\STATE $\boldsymbol{\gamma}_i^\beta =\gamma_i\frac{\boldsymbol{\zeta}_{i+1}^\beta}{\boldsymbol{\zeta}_{i}^\beta}$
		\STATE $W_{i+1}^\beta=W_i^\beta-\boldsymbol{\gamma}_i^\beta P_i^\beta$}
		\STATE $\textbf{r}_{i+1}=\textbf{r}_i+\gamma_i\textbf{q}_i$
		\IF{$||\textbf{r}_{i+1}||<tol$} 
		\STATE Break
		\ENDIF 
		\STATE $\alpha_{i+1}=\frac{\langle\textbf{r}_{i+1},\textbf{r}_{i+1}\rangle}{\langle\textbf{r}_i,\textbf{r}_i\rangle}$
		\STATE $\textbf{p}_{i+1}=\textbf{r}_{i+1}+\alpha_{i+1}\textbf{p}_i$
		{\color{blue!60!black}\STATE$\boldsymbol{\alpha}_{i+1}^\beta=\alpha_i\frac{\boldsymbol{\zeta}_{i+1}^\beta\boldsymbol{\gamma}_{i}^\beta}{\boldsymbol{\zeta}_{i}^\beta\gamma_i}$
		\STATE $P_{i+1}^\beta=\boldsymbol{\zeta}_{i+1}^\beta\textbf{r}_{i+1}+\boldsymbol{\alpha}_{{i+1}}^\beta P_i^\beta$ 
		\FOR{$s=0,\dots, \text{size}(\boldsymbol{\beta})$} 
			\IF{$||\boldsymbol{\zeta}_{i+1}^\beta(s)\textbf{r}_{i+1}||<tol$} 
			\STATE Stop updates for $\boldsymbol{\beta}(s)$
			\ENDIF  
			\ENDFOR}
		\ENDFOR
	\end{algorithmic}
\end{algorithm} 
\end{minipage}
\caption{The Conjugate Gradient algorithm in Algorithm \ref{alg:cg} \cite{demmel1997applied} and the Conjugate Gradient algorithm for various shifted linear systems in Algorithm \ref{alg:Shifted_cg} \cite{jegerlehner1996krylov,sciarracgalgorithm}. The additional operations for shifted CG are highlighted in blue.}
\end{figure*}

\subsubsection{Centralization for a Krylov subspace}
In Section \ref{sec:centralization} it was shown that centralization avoids calculating the non-discriminative eigenvector. Note that if we apply centralization directly on the matrix $X$, we will lose the sparsity of the data matrix and one of the advantages of Krylov subspace methods. In order to avoid creating a dense matrix filled with real values, the centralization is applied in the matrix-vector product. Using $\textbf{z}_{nd}=\textbf{1}\in\mathbb{R}^{N\times 1}$ as the vector of all ones and defining $\textbf{z}_\ell=\textbf{1}_\ell\in\mathbb{R}^{N\times 1}$ as the vector with value $1$ for the labeled samples and $0$ otherwise, the mean labeled sample in the feature space $\boldsymbol{\mu}_\ell=X^T\textbf{1}_\ell/\ell$ can then be derived. The centralized multiplication with $X$ can be written as
\begin{align*}\left[(I_N-\frac{\textbf{1}\textbf{1}_\ell^T}{\ell})X\right]\textbf{v}&=(X-\textbf{1}\boldsymbol{\mu}_\ell^T)\textbf{v}\\ &=X\textbf{v}-\textbf{1}\langle \boldsymbol{\mu}_\ell^T,\textbf{v}\rangle\end{align*}  and the centralized multiplication with $X^T$ as \begin{align*}\left[(I_N-\frac{\textbf{1}\textbf{1}_\ell^T}{\ell})X\right]^T\textbf{w}&=(X-\textbf{1}\boldsymbol{\mu}_\ell^T)^T\textbf{w}\\&=X^T\textbf{w}-\boldsymbol{\mu}_\ell\langle\textbf{1}^T,\textbf{w}\rangle\end{align*} were the last inner product is the sum of the elements of $\textbf{w}$.   

For the SDA eigenvalue problem, we only have to implement one centralized multiplication. For example, using the centralized multiplication with $X^T$, the matrix-vector product in SDA for the non-discriminative eigenvector becomes 
\begin{align*}
\left(\left[(I_N-\frac{\textbf{1}\textbf{1}_\ell^T}{\ell})X\right]^T\left[(1-\alpha)\begin{pmatrix}
I_\ell&0\\0&0\end{pmatrix}+\alpha L\right] X\right)\textbf{w}_{nd}\\=\left[(I_N-\frac{\textbf{1}\textbf{1}_\ell^T}{\ell})X\right]^T\left[(1-\alpha)\begin{pmatrix}
I_\ell&0\\0&0\end{pmatrix}+\alpha L\right] \textbf{z}_{nd}\\
=\left[(I_N-\frac{\textbf{1}\textbf{1}_\ell^T}{\ell})X\right]^T(1-\alpha) \textbf{z}_\ell\\
=X^T(I_N-\frac{\textbf{1}_\ell\textbf{1}^T}{\ell})(1-\alpha)\textbf{z}_\ell\\
=(1-\alpha)X^T(\textbf{z}_\ell-\frac{\ell\textbf{z}_\ell}{\ell})\\
=\textbf{0}.
\end{align*}
Using the centralized multiplication with $X^T$ avoids the non-discriminative eigenvector. This holds for the centralized multiplication with $X$ as well and the derivation is similar. 

\section{Four algorithms for SDA}\label{sec:proposed}
Here we present four methods for solving the SDA eigenvalue problem for binary classification. First, we propose an efficient implementation of the spectral regression for SDA (SR-SDA). Next, we present our improved version of the spectral regression by adding centralization to SR-SDA, resulting in centralized spectral regression for SDA (CSR-SDA). We then propose spectral analysis as a standalone method to classify the data (SA-SDA) in the semi-supervised setting. Finally, we present our scalable implementation for solving the centralized SDA eigenvalue problem directly (FSDA). The different algorithms are implemented in C++.  

For the evaluation of the algorithms, we use area under the curve of the receiver operating characteristic (AUC-ROC). In the case of binary classification, there is only one discriminative eigenvector. To perform the classification using this eigenvector, the data is projected along this vector, namely $X\textbf{w}=\textbf{s}\in\mathbb{R}^{N\times 1}$, and results in the sample rating or score. Here we use the rating directly to calculate the AUC-ROC, but other classification methods like support vector machines \cite{Vapnik1999overview}, nearest neighbor \cite{Cover1967Nearest} or nearest centroid \cite{McIntyre1980Centroid,Tibshirani2002Diagnosis} can be used to classify the data based on the projection.
\subsection{Spectral regression for SDA}
For the spectral regression framework applied to SDA (\sr), we need to solve the spectral analysis SDA eigenvalue problem (\ref{eqn:sasdaeig}). The original framework finds all the eigenvectors of the spectral analysis SDA eigenvalue problem and then finds the corresponding feature vectors by solving a linear system in a regularized least squares manner. In order to find the eigenvectors, we use subspace iteration with the block conjugate gradient method (BCG) \cite{o1980block} as the linear system solver. Subspace iteration will actually find the eigenspace $Z\in\mathbb{R}^{N\times 2}$ containing both the non-discriminative eigenvector $\textbf{z}_{nd}$ and the discriminative eigenvector $\textbf{z}_{d}$. SR would normally proceed by computing the corresponding vectors $\textbf{w}_{nd}$ and $\textbf{w}_{d}$ in the regression phase. The computed space $Z$ during the spectral analysis phase consists of linear combinations of $\textbf{z}_{nd}$ and $\textbf{z}_{d}$. To compare the original SR with our improvements, we need the discriminative eigenvector to retrieve the rating of the data samples and compute the AUC-ROC. 

The discriminative eigenvector can be extracted by applying Rayleigh-Ritz subspace projection \cite{parlett1998symmetric}. Using the eigenspace $Z$, eigenvalue problem (\ref{eqn:sasdaeig}) is transformed to a two-by-two eigenvalue problem $Z^TAZ\textbf{q}_i=\lambda_iZ^TBZ\textbf{q}_i$ with $i=1,2$ and $A$, $B$ shorthands. The discriminative eigenvector is then the projection of the second largest eigenvector $\textbf{z}_d=Z\textbf{q}_2$. Next we solve the linear system $(X^TX+\beta I)\textbf{w}_d=X^T\textbf{z}_d$ using shifted CG (Algorithm \ref{alg:Shifted_cg}) to find the discriminative eigenvector $\textbf{w}_d$ in the feature space. The full algorithm is given in Algorithm \ref{alg:SR}, where we apply the regression phase to both eigenvectors  $\textbf{z}_d$ and  $\textbf{z}_{nd}$ to match the original algorithm as close as possible. The LSQR method is advocated for linear systems arising from regularized least squares \cite{paige1982lsqr} and used in spectral regression. However, to the best of our knowledge, there is no specific LSQR variant exploiting the shift-invariance property of Krylov subspaces \footnote{The spectral regression code \cite{cai2007spectral} of the authors is available at \url{http://www.cad.zju.edu.cn/home/dengcai/Data/SR.html} and is for MATLAB which uses the function \textit{eigs}. Solving the spectral analysis eigenvalue problem for SDA using a toy data set with \textit{eigs} already took twice as long as solving the spectral SDA eigenvalue problem and the linear system with our own C++ code.}. As mentioned, our implementation thus uses shifted CG to solve the system for multiple regularization parameters.
\begin{algorithm*}
	\caption{\sr}
	\label{alg:SR}
	\begin{algorithmic}[1]
		\STATE Given $\alpha$, $\beta$
		\STATE Initialize $R\in \mathbb{R}^{N\times 2}$ randomly generated from a uniform distribution between $-1$ and $1$
		\STATE Solve $Z\in \mathbb{R}^{N\times 2}$ with BCG from $$\left[(1-\alpha)\begin{pmatrix}
		I_\ell&0\\0&0\end{pmatrix}+\alpha L\right]Z= \begin{pmatrix}
		W & 0\\0&0
		\end{pmatrix}R$$
		\STATE Solve two-by-two eigenvalue problem $$ Z^T\begin{pmatrix}
		W & 0\\0&0
		\end{pmatrix}Z\textbf{q}_i=\lambda_i Z^T\left[(1-\alpha)\begin{pmatrix}
		I_\ell&0\\0&0\end{pmatrix}+\alpha L\right]Zq_i$$
		\FOR{$i=1,2$}
		\STATE Solve $\textbf{w}_i$ with shifted CG from $$(X^TX+\beta I)\textbf{w}_i=X^T(Z\textbf{q}_i)$$
		\ENDFOR
		\RETURN Rating: $\textbf{s}=X\textbf{w}_2$
	\end{algorithmic}
\end{algorithm*} 
\subsection{Centralized spectral regression for SDA}
SR as detailed above computes the redundant non-discriminative eigenvector $\textbf{z}_{nd}$ which can be avoided by looking at the SA phase in SDA (\ref{eqn:sasdaeig}). The SA phase is actually an SDA eigenvalue problem (\ref{eqn:SDAeig}) as well with the identity matrix $I_N$ as the data matrix $X$ . In order to avoid the non-discriminative eigenvector $\textbf{z}_{nd}$, we centralize the identity matrix around the mean of the labeled data ($\textbf{1}_\ell/\ell$). Now we only need to compute one eigenvector and therefore using the power method \cite{parlett1998symmetric} suffices, which is basically subspace iteration with only one right-hand side. The difference with \sr\ is that CSR-SDA computes one eigenvector less during the spectral analysis phase and therefore requires only one linear solve during the regression phase. The full algorithm for centralized spectral regression for SDA (CSR-SDA) is given is Algorithm \ref{alg:optimizedSR}. 
\begin{algorithm*}
	\caption{\osr}
	\label{alg:optimizedSR}
	\begin{algorithmic}[1]
		\STATE Given $\alpha$, $\beta$
		\STATE Initialize $\textbf{r}\in \mathbb{R}^{N\times 1}$ randomly generated from a uniform distribution between $-1$ and $1$
		\STATE Orthogonalize: $\textbf{r}=\textbf{r} -\textbf{1}\frac{<\textbf{1}_\ell^T,\textbf{r}>}{\ell}$
		\STATE Solve $\textbf{z}$ with CG from $$\left((I_N-\frac{\textbf{1}\textbf{1}_\ell^T}{\ell})^T\left[(1-\alpha)\begin{pmatrix}
		I_\ell&0\\0&0\end{pmatrix}+\alpha L\right] I_N\right)\textbf{z}= \begin{pmatrix}
		W & 0\\0&0
		\end{pmatrix}\textbf{r}$$
		\STATE Solve $\textbf{w}$ with shifted CG from $$(X^TX+\beta I)\textbf{w}=X^T\textbf{z} $$
		\RETURN Rating: $\textbf{s}=X\textbf{w}$
	\end{algorithmic}
\end{algorithm*} 
\subsection{Spectral analysis for SDA}
In the semi-supervised setting, the data samples requiring classification are often available as unlabeled data. The regression step in CSR-SDA solves the linear system $\textbf{z}=X\textbf{w}$ in a regularized least-squares manner (Algorithm \ref{alg:optimizedSR}, line 5). The rating of the samples is next calculated as the projection of the data point along the discriminative direction $\textbf{s}=X\textbf{w}$. For the rating of the known unlabeled data the eigenvector $\textbf{z}$ can be used ($\textbf{s}=\textbf{z}$), making the regression step obsolete as shown in Algorithm \ref{alg:sasda}. We added regularization to the SA phase as countermeasure against overfitting. Spectral analysis for SDA (SA-SDA) as standalone method only works when the samples to be classified are known and present in $X$. If $\alpha=0$, the Laplacian is neutralized and the unlabeled data receive a rating of zero. Therefore, this approach only works if $\alpha \ne 0$. 
\begin{algorithm*}
	\caption{\ndsr}
	\label{alg:sasda}
	\begin{algorithmic}[1]
		\STATE Given $\alpha$, $\beta$
		\STATE Initialize $\textbf{r}\in \mathbb{R}^{N\times 1}$ randomly generated from a uniform distribution between $-1$ and $1$
		\STATE Orthogonalize: $\textbf{r}=\textbf{r} -\textbf{1}\frac{<\textbf{1}_\ell^T,\textbf{r}>}{\ell}$
		\STATE Solve $\textbf{z}$ with shifted CG from $$\left((I_N-\frac{\textbf{1}\textbf{1}_\ell^T}{\ell})^T\left[(1-\alpha)\begin{pmatrix}
		I_\ell&0\\0&0\end{pmatrix}+\alpha L\right] I_N+\beta I_N\right)\textbf{z}= \begin{pmatrix}
		W & 0\\0&0
		\end{pmatrix}\textbf{r}$$
		\RETURN Rating: $\textbf{s}=\textbf{z}$
	\end{algorithmic}
\end{algorithm*}   
\subsection{Fast semi-supervised discriminant analysis}
Since we are working with sparse data and Krylov subspace methods, the complexity of the algorithms are proportional to the number of nonzeros. Hence, we propose an efficient implementation to solve the centralized SDA eigenvalue problem directly by using the power method. The full algorithm for fast semi-supervised discriminant analysis (FSDA) is given in Algorithm \ref{alg:ourmethod}.
\begin{algorithm*}
	\caption{\algname}
	\label{alg:ourmethod}
	\begin{algorithmic}[1]
		\STATE Given $\alpha$, $\beta$
		\STATE Initialize $\textbf{r}\in \mathbb{R}^{D\times 1}$ randomly generated from a uniform distribution between $-1$ and $1$
		\STATE Calculate mean labeled sample: $\ell\boldsymbol{\mu}_\ell=X^T\textbf{1}_\ell$
		\STATE Solve $\textbf{w}$ with shifted CG from $$\left((X-\textbf{1}\boldsymbol{\mu}_\ell^T)^T\left[(1-\alpha)\begin{pmatrix}
		I_\ell&0\\0&0\end{pmatrix}+\alpha L\right] X+\beta I_D\right)\textbf{w}= (X-\textbf{1}\boldsymbol{\mu}_\ell^T)^T\begin{pmatrix}
		W & 0\\0&0
		\end{pmatrix}X\textbf{r}$$
		\RETURN Rating: $\textbf{s}=X\textbf{w}$
	\end{algorithmic}
\end{algorithm*}  
\subsection{Comparison of the SDA algorithms}
All four algorithms solve the same underlying problem. In a perfect world without regularization and if the solution of $\textbf{z}_d=X\textbf{w}_d$ exists, these algorithms would have the same performance. In a real world regularization is required and iterative methods are used to solve linear systems within a certain tolerance. The difference between \sr\ and \osr\ is only computational and these two algorithms result in the same classification. The differences between \osr, \ndsr\ and \algname\ are more subtle. \algname\ solves the SDA eigenvalue problem (\ref{eqn:SDAeig}) directly and the tolerance used in CG reflects the true error of the eigenvector. For \osr\ using the same tolerance in both linear systems does not guarantee this tolerance for the SDA eigenvalue problem.     

All presented iterative methods build a Krylov subspace and each iteration is linear in the number of samples, features and nonzeros of the matrices. Furthermore, centralization costs $4$ linear operations for each iteration with linear the column size of the matrix used. Table \ref{tab:sdacomplexity} shows the complexity of the different algorithms given in this section. Applying centralization for \osr\ results in a speed-up of approximately $C/(C-1)$ with respect to \sr, reducing execution time by almost a factor $2$ for binary classification.

The table also shows the main difference between CG and Shifted CG. We did not take into account the influence of the $\beta$-value on the number of iterations required to reach a certain tolerance and assumed that each solve takes the same number of iterations. CG has to rebuild the Krylov subspace for each value of $\beta$, while shifted CG reuses the base resulting in additional linear operations. Note that the number of nonzeros $nz$ is generally significantly larger than the column size of the matrix, and thus shifted CG results in a speed-up approximately the number of regularization parameters used. 
\begin{table}[ht]
	\renewcommand{\arraystretch}{1.0}
	\begin{center}
		\scalebox{1.0}{
			\begin{tabular}{l|p{5.5cm}}\toprule
				Algorithm&Complexity\\ \midrule
				\sr&$k_1C(nz_L+10N)+k_2C(2nz_X+10D)$\\
				\osr&$k_1(C-1)(nz_L+14N)+k_2(C-1)(2nz_X+10D)$\\
				\ndsr&$k_1(C-1)(nz_L+14N)$\\
				\algname&$k_2(C-1)(2nz_X+nz_L+14D)$\\
				CG & $k_1N_\beta(nz+10N)$\\
				Shifted CG&  $k_1(nz+N_\beta10N+P_\beta)$\\
			  \bottomrule
				\multicolumn{2}{l}{with}\\
				\multicolumn{2}{l}{$k_1$: the number of CG-iterations with dimension $N$}\\
				\multicolumn{2}{l}{$k_2$: the number of CG-iterations with dimension $D$}\\
				\multicolumn{2}{l}{$nz_L$: the number of nonzeros of $L$}\\
				\multicolumn{2}{l}{$nz_X$: the number of nonzeros of $X$}\\
				\multicolumn{2}{l}{$N_\beta$: the number regularization parameters $\beta$}\\
				\multicolumn{2}{l}{$P_\beta$: extra cost related to $N_\beta$ used in shifted CG}\\
		\end{tabular}}
	\end{center}
	\caption{The complexity of the 4 SDA algorithms.}\label{tab:sdacomplexity}
\end{table}

\section{Experimental results}\label{sec:results}
For evaluating the different methods, we present the AUC-ROC in function of the execution time for the data sets. Since these methods solve the same underlying problem, we are interested in convergence speed and difference in the algorithms predictive performances.    

This work is performed with drug-protein activity prediction in mind. We have tested our methods on different targets for two different data sets. The first data set is described in section \ref{subsect:ChEMBL} and is publicly available. The implementation for this data set was compiled with gcc 5.3.1 and OPENMP 4.0 with compile option -O3 and executed on a machine with an Intel(R) Core(TM) i7-6560U (2.20GHz) processor with an L3 cache memory of 4096 Kb and 16 GB of DRAM where all 4 cores were used \footnote{Implementation and the public ChEMBL data set can be found here: \url{https://people.cs.kuleuven.be/~joris.tavernier/}.}. The second industry-scale data set is described in section \ref{subsect:Janssen} and stronger hardware was required. These experiments were run using a computing node with dual Intel(R) Xeon(R) E5-2699 v3 (2.30GHz) processors with a L3 cache memory of 46080 Kb and 264 gb of DRAM. For these experiments we used 12 cores.  
\subsection{ChEMBL data}\label{subsect:ChEMBL}
For drug-protein activity prediction, one of the possible indicators is the IC50 value. IC50 or the half maximal inhibitory concentration measures the substance concentration required to inhibit the activity of a protein by $50\%$. Using the bioactivity database ChEMBL version 19 \cite{bento2014chembl}, we selected protein targets that had at least 200 or more IC50 values for the compounds. A compound was considered active with an IC50 value less than 1000 nM. We then kept the protein targets that had at least 30 actives and inactives, resulting in 330 targets with enough labels to perform 5-fold nested cross-validation. 

\begin{figure*}[ht]
	\begin{subfigure}{0.37\textwidth}
		\tikzsetnextfilename{ocstruc2}
		\centering
		\resizebox{0.5\linewidth}{!}{\begin{tikzpicture}
\node (A) at (2,0) {\chemfig{H-[:0]C (-[:270]O-[:270]H)(-[:90]H)-[:0]C(-[270]O-[:270]H)(-[:0]H)(-[:90]H)} };
\node (A) at (2,2) {};
\node (A) at (2,-2) {};
\end{tikzpicture}   

%\node (A) {\chemfig{C(-[:0]H)(-[:90]H)(-[:180]H)(-[:270]H)} };
%\chemfig{A-[:30]B=[:-75]C-[:10]D-[:90]>|[:60]-[:-20]E-[:0]~[:-75]F}
}
		\caption{Compound structure}
		\label{fig:ocstruc2}
	\end{subfigure}
	\begin{subfigure}{0.59\textwidth}
		\tikzsetnextfilename{octree2}
		\centering
		\resizebox{\linewidth}{!}{\begin{tikzpicture}
\node (O1) at (0,0) {O };
\node (C1) at (2,0) {C };
\node (O2) at (0,-2) {O };
\node (C2) at (2,-2) {C };
\node (OC2) at (4,-2) {[O,1,C] };
\node (COC2) at (6,-2) {[C,1,O,1,C] };
\node (O3) at (0,-4) {O };
\node (C3) at (2,-4) {C };
\node (OC3) at (4,-4) {[O,1,C] };
\node (COC3) at (6,-4) {[C,1,O,1,C] };
\node (OCCO3) at (9.5,-4) {[[O,1,C],1,[C,1,C,1,C,1,O]] };
\node (l1) at (12.2,0) {Layer 1 };
\node (l2) at (12.2,-2) {Layer 2 };
\node (l3) at (12.2,-4) {Layer 3 };
\draw[->,gray,label = 1] (O1) edge (O2);
\draw[->,gray,label = 1] (O2) edge (O3);
\draw[->,gray,label = 1] (O1) edge (OC2);
\draw[->,gray,label = 1] (C1) edge (OC2);
\draw[->,gray,label = 1] (OC2) edge (OC3);
\draw[->,gray,label = 1] (C1) edge (C2);
\draw[->,gray,label = 1] (O1) edge (COC2);
\draw[->,gray,label = 1] (C1) edge (COC2);
\draw[->,gray,label = 1] (COC2) edge (COC3);
\draw[->,gray,label = 1] (C2) edge (C3);
\draw[->,gray,label = 1] (OC2) edge (OCCO3);
\draw[->,gray,label = 1] (COC2) edge (OCCO3);
\end{tikzpicture}   

%\node (A) {\chemfig{C(-[:0]H)(-[:90]H)(-[:180]H)(-[:270]H)} };
%\chemfig{A-[:30]B=[:-75]C-[:10]D-[:90]>|[:60]-[:-20]E-[:0]~[:-75]F}
}
		\caption{Resulting features}
		\label{fig:octree2}
	\end{subfigure}
	\caption{Example of the ECFP feature generation.}
	\label{fig:ecfpexample}
\end{figure*}
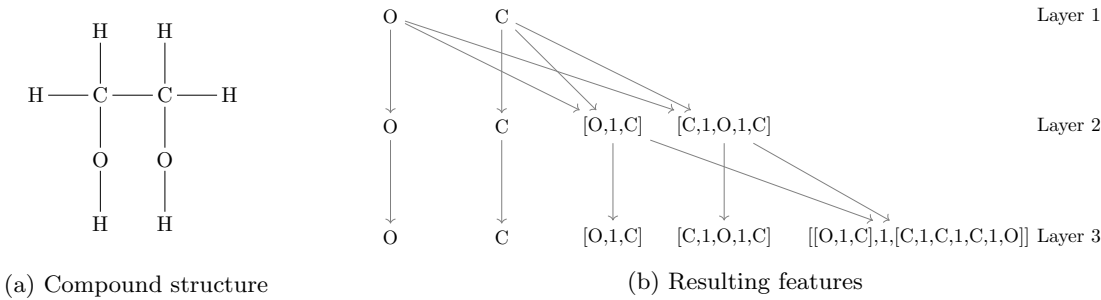

For the compounds the extended-connectivity fingerprints (ECFP \cite{rogers2010extended}) were computed using rdkit \cite{rdkit} with 3 layers. In summary, ECFP generates the features for a compound in layers. In the first layer each individual atom receives an initial identifier based on its atomic symbol. In the following layer this value is concatenated with the bound and the identifiers of the neighboring atoms and then hashed to a new identifier for this atom. This process is repeated for a fixed number of layers.  These generated substructures represented by their hash value are the features for this compound. Duplicate identifiers or duplicate substructures are removed from the feature list. Figure \ref{fig:ecfpexample} depicts the creation of the ECFP features for a small chemical compound. The characteristics of the resulting ECFP data matrix $X$ are given in Table \ref{tab:Chemblcharac}. 
\begin{table}[ht]
	\renewcommand{\arraystretch}{1.0}
	\begin{center}
		\scalebox{1.0}{
			\begin{tabular}{l|c}\toprule
				Parameter&Value\\\midrule
				Compounds&$167\ 668$\\
				Features&$291\ 714$\\
				Nonzeros&$12\ 246\ 376$\\
				Density&$0.025038\%$ \\\bottomrule
			\end{tabular}}
		\end{center}
		\caption{The characteristics of the ChEMBL data matrix.}\label{tab:Chemblcharac}
	\end{table}
	
The similarity matrix was made using a $5$-nearest neighbor graph and the Tanimoto similarity. The cross-validated mean and standard deviation for AUC-ROC of FSDA over the 330 protein targets is $89 \pm 4\%$. In addition, to evaluate the convergence and execution time for the different methods, we have randomly selected 6 targets with 2000 or more labeled compounds. The number of active and inactive compounds for each target is given in Table \ref{tab:chembltargets}.   

\begin{table*}[ht]
	\renewcommand{\arraystretch}{1.0}
	\begin{center}
		\scalebox{1.0}{
			\begin{tabular}{l|c|c|c|c} \toprule
				Name& Active& Inactive& Known labels&$\%$ labeled\\ \midrule
				CHEMBL100503& 2225& 1729 &3954& 2.36\\	
				CHEMBL10062& 1365&1275 &2640&1.57\\ 		
				CHEMBL100670& 1621&491 &2112&1.26\\ 						
				CHEMBL1007& 1409& 1170& 2579&1.54\\ 		
				CHEMBL100763& 817& 3144&3961&2.36\\ 		
				CHEMBL100875& 2514& 516&	3030& 1.81\\ \bottomrule
			\end{tabular}}
		\end{center}
		\caption{The characteristics of the chosen targets for the 167668 chemical compounds of the ChEMBL data set.}\label{tab:chembltargets}
	\end{table*} 

For all methods, we varied the number of CG iterations which increases the complexity, but also the accuracy of CG. Note that the spectral regression methods require two linear solves and the best pair (execution time, AUC-ROC) is reported. The different number of iterations for the linear solvers are given in Table \ref{tab:exp}. 5-Fold nested cross-validation was used for the experiments, with the inner-folds used to pick the best value of the regularization parameter $\beta$. Starting from $\beta=10^{-9}$, we let $\beta$ vary by multiplying $\beta$ with $10$ until $\beta=10^{3}$. We report the average execution time for the outer cross-validation using only one optimal $\beta$. The reported AUC-ROC is the average along with the standard deviation for all outer folds and 5 different seed initializations of the random number generator. 
\begin{table*}[ht]
	\renewcommand{\arraystretch}{1.0}
	\begin{center}
		\scalebox{1.0}{
			\begin{tabular}{l|p{7cm}|l}\toprule
				Method& Description& Iterations\\ \midrule
				\algname&Iterations linear solve for SDA & $[2,3,5,10,20,40,60,80]$\\ \midrule
				\osr&Iterations linear solve for spectral SDA &$[2,3,5,10,20,40,60,80]$ \\
				&Iterations RLS solve with $X$ (regression) & $[2,3,5,10,20,40,60,80]$ \\ \midrule
				\ndsr&Iterations linear solve for spectral SDA &$[3,5,10,20,40,60,80]$ \\ \midrule
				\sr &Iterations linear solve for spectral SDA &$[2,3,5,10,20,40,60,80]$ \\
				&Iterations RLS solve with $X$ (regression) & $[2,3,5,10,20,40,60,80]$ \\\bottomrule
			\end{tabular}}
		\end{center}
		\caption{The different parameters used in the experiments for each method.}\label{tab:exp}
	\end{table*}

\begin{figure*}
	\begin{subfigure}[b]{0.46\textwidth}
		\tikzsetnextfilename{A}
		\centering
		\resizebox{\linewidth}{!}{\begin{tikzpicture}
\begin{axis}[
height=8cm,
width=12cm,
xmode = log,
cycle list name=colors,
%ymode = log,
grid=both,
grid style={line width=.1pt, draw=gray!20},
major grid style={line width=.2pt,draw=gray!60},
minor tick num=5,
xmin=0.02,xmax=6,
xlabel=Time(s),ylabel=AUC-ROC,
legend pos=south east,
legend style={font=\small}
]
\addplot+[mark=none,very thick,solid,error bars/.cd,y dir=both,y explicit,x dir=both,x explicit,error bar style={solid,very thin}] table[x=time,x error= std-time, y=auc,y error=std-auc] {Tikz/gene79_Subspace.dat};

\addplot+[mark=none,very thick,densely dotted,error bars/.cd,y dir=both,y explicit,x dir=both,x explicit,error bar style={solid,very thin}] table[x=time,x error= std-time, y=auc,y error=std-auc] {Tikz/gene79_Sr_o.dat};
\addplot+[mark=none,dashed,very thick,error bars/.cd,y dir=both,y explicit,x dir=both,x explicit,error bar style={solid,very thin}] table[x=time,x error= std-time, y=auc,y error=std-auc] {Tikz/gene79_Sr_nd.dat};
\addplot+[mark=none,very thick,dashdotted,error bars/.cd,y dir=both,y explicit,x dir=both,x explicit,error bar style={solid,very thin}] table[x=time,x error= std-time, y=auc,y error=std-auc] {Tikz/gene79_Sr.dat};
\legend{\algname\ (ours), \osr\ (ours),\ndsr\ (ours), \sr}
\end{axis}
\end{tikzpicture}   }
		\caption{Target CHEMBL100503}
		\label{fig:A}
	\end{subfigure}
	\begin{subfigure}[b]{0.46\textwidth}
		\tikzsetnextfilename{B}
		\centering
		\resizebox{\linewidth}{!}{\begin{tikzpicture}
\begin{axis}[
height=8cm,
width=12cm,
xmode = log,
cycle list name=colors,
%ymode = log,
grid=both,
grid style={line width=.1pt, draw=gray!20},
major grid style={line width=.2pt,draw=gray!60},
minor tick num=5,
xlabel=Time(s),ylabel=AUC-ROC,
legend pos=south east,
legend style={font=\small}
]
\addplot+[mark=none,very thick,solid,error bars/.cd,y dir=both,y explicit,x dir=both,x explicit,error bar style={solid,very thin}] table[x=time,x error= std-time, y=auc,y error=std-auc] {Tikz/gene102_Subspace.dat};
\addplot+[mark=none,very thick,densely dotted,error bars/.cd,y dir=both,y explicit,x dir=both,x explicit,error bar style={solid,very thin}] table[x=time,x error= std-time, y=auc,y error=std-auc] {Tikz/gene102_Sr_o.dat};
\addplot+[mark=none,dashed,very thick,error bars/.cd,y dir=both,y explicit,x dir=both,x explicit,error bar style={solid,very thin}] table[x=time,x error= std-time, y=auc,y error=std-auc] {Tikz/gene102_Sr_nd.dat};
\addplot+[mark=none,very thick,dashdotted,error bars/.cd,y dir=both,y explicit,x dir=both,x explicit,error bar style={solid,very thin}] table[x=time,x error= std-time, y=auc,y error=std-auc] {Tikz/gene102_Sr.dat};
\legend{\algname\ (ours), \osr\ (ours),\ndsr\ (ours), \sr}
\end{axis}
\end{tikzpicture}   } 
		\caption{Target CHEMBL10062}
		\label{fig:B}
	\end{subfigure}
	
	\begin{subfigure}[b]{0.46\textwidth}
		\tikzsetnextfilename{C}
		\centering
		\resizebox{\linewidth}{!}{\begin{tikzpicture}
\begin{axis}[
height=8cm,
width=12cm,
xmode = log,
cycle list name=colors,
%ymode = log,
grid=both,
grid style={line width=.1pt, draw=gray!20},
major grid style={line width=.2pt,draw=gray!60},
minor tick num=5,
xmin=0.02,xmax=6,
xlabel=Time(s),ylabel=AUC-ROC,
legend pos=south east,
legend style={font=\small}
]
\addplot+[mark=none,very thick,solid,error bars/.cd,y dir=both,y explicit,x dir=both,x explicit,error bar style={solid,very thin}] table[x=time,x error= std-time, y=auc,y error=std-auc] {Tikz/gene112_Subspace.dat};

\addplot+[mark=none,very thick,densely dotted,error bars/.cd,y dir=both,y explicit,x dir=both,x explicit,error bar style={solid,very thin}] table[x=time,x error= std-time, y=auc,y error=std-auc] {Tikz/gene112_Sr_o.dat};
\addplot+[mark=none,dashed,very thick,error bars/.cd,y dir=both,y explicit,x dir=both,x explicit,error bar style={solid,very thin}] table[x=time,x error= std-time, y=auc,y error=std-auc] {Tikz/gene112_Sr_nd.dat};
\addplot+[mark=none,very thick,dashdotted,error bars/.cd,y dir=both,y explicit,x dir=both,x explicit,error bar style={solid,very thin}] table[x=time,x error= std-time, y=auc,y error=std-auc] {Tikz/gene112_Sr.dat};
\legend{\algname\ (ours), \osr\ (ours),\ndsr\ (ours), \sr}
\end{axis}
\end{tikzpicture}     }
		\caption{Target CHEMBL100670}
		\label{fig:C}
	\end{subfigure}
	\begin{subfigure}[b]{0.46\textwidth}
		\tikzsetnextfilename{A2}
		\centering
		\resizebox{\linewidth}{!}{\begin{tikzpicture}
\begin{axis}[
height=8cm,
width=12cm,
cycle list name=colors,
xmode = log,
%ymode = log,
grid=both,
grid style={line width=.1pt, draw=gray!20},
major grid style={line width=.2pt,draw=gray!60},
minor tick num=5,
xmin=0.02,xmax=6,
xlabel=Time(s),ylabel=AUC-ROC,
legend pos=south east,
legend style={font=\small}
]
\addplot+[mark=none,very thick,solid,error bars/.cd,y dir=both,y explicit,x dir=both,x explicit,error bar style={solid,very thin}] table[x=time,x error= std-time, y=auc,y error=std-auc] {Tikz/gene115_Subspace.dat};
\addplot+[mark=none,very thick,densely dotted,error bars/.cd,y dir=both,y explicit,x dir=both,x explicit,error bar style={solid,very thin}] table[x=time,x error= std-time, y=auc,y error=std-auc] {Tikz/gene115_Sr_o.dat};
\addplot+[mark=none,dashed,very thick,error bars/.cd,y dir=both,y explicit,x dir=both,x explicit,error bar style={solid,very thin}] table[x=time,x error= std-time, y=auc,y error=std-auc] {Tikz/gene115_Sr_nd.dat};
\addplot+[mark=none,very thick,dashdotted,error bars/.cd,y dir=both,y explicit,x dir=both,x explicit,error bar style={solid,very thin}] table[x=time,x error= std-time, y=auc,y error=std-auc] {Tikz/gene115_Sr.dat};
\legend{\algname\ (ours), \osr\ (ours),\ndsr\ (ours), \sr}
\end{axis}
\end{tikzpicture}   }
		\caption{Target CHEMBL1007}
		\label{fig:A2}
	\end{subfigure}
	
	\begin{subfigure}[b]{0.46\textwidth}
		\tikzsetnextfilename{B2}
		\centering
		\resizebox{\linewidth}{!}{\begin{tikzpicture}
\begin{axis}[
height=8cm,
width=12cm,
xmode = log,
cycle list name=colors,
%ymode = log,
grid=both,
grid style={line width=.1pt, draw=gray!20},
major grid style={line width=.2pt,draw=gray!60},
minor tick num=5,
xmin=0.02,xmax=6,
xlabel=Time(s),ylabel=AUC-ROC,
legend pos=south east,
legend style={font=\small}
]
\addplot+[mark=none,very thick,solid,error bars/.cd,y dir=both,y explicit,x dir=both,x explicit,error bar style={solid,very thin}] table[x=time,x error= std-time, y=auc,y error=std-auc] {Tikz/gene129_Subspace.dat};

\addplot+[mark=none,very thick,densely dotted,error bars/.cd,y dir=both,y explicit,x dir=both,x explicit,error bar style={solid,very thin}] table[x=time,x error= std-time, y=auc,y error=std-auc] {Tikz/gene129_Sr_o.dat};
\addplot+[mark=none,dashed,very thick,error bars/.cd,y dir=both,y explicit,x dir=both,x explicit,error bar style={solid,very thin}] table[x=time,x error= std-time, y=auc,y error=std-auc] {Tikz/gene129_Sr_nd.dat};
\addplot+[mark=none,very thick,dashdotted,error bars/.cd,y dir=both,y explicit,x dir=both,x explicit,error bar style={solid,very thin}] table[x=time,x error= std-time, y=auc,y error=std-auc]  {Tikz/gene129_Sr.dat};
\legend{\algname\ (ours), \osr\ (ours),\ndsr\ (ours), \sr}
\end{axis}
\end{tikzpicture}   } 
		\caption{Target CHEMBL100763}
		\label{fig:B2}
	\end{subfigure}
	\begin{subfigure}[b]{0.46\textwidth}
		\tikzsetnextfilename{C2}
		\centering
		\resizebox{\linewidth}{!}{\begin{tikzpicture}
\begin{axis}[
height=8cm,
width=12cm,
xmode = log,
cycle list name=colors,
%ymode = log,
grid=both,
grid style={line width=.1pt, draw=gray!20},
major grid style={line width=.2pt,draw=gray!60},
minor tick num=5,
xmin=0.02,xmax=6,
xlabel=Time(s),ylabel=AUC-ROC,
legend pos=south east,
legend style={font=\small}
]
\addplot+[mark=none,very thick,solid,error bars/.cd,y dir=both,y explicit,x dir=both,x explicit,error bar style={solid,very thin}] table[x=time,x error= std-time, y=auc,y error=std-auc] {Tikz/gene155_Subspace.dat};
\addplot+[mark=none,very thick,densely dotted,error bars/.cd,y dir=both,y explicit,x dir=both,x explicit,error bar style={solid,very thin}] table[x=time,x error= std-time, y=auc,y error=std-auc] {Tikz/gene155_Sr_o.dat};
\addplot+[mark=none,dashed,very thick,error bars/.cd,y dir=both,y explicit,x dir=both,x explicit,error bar style={solid,very thin}] table[x=time,x error= std-time, y=auc,y error=std-auc] {Tikz/gene155_Sr_nd.dat};
\addplot+[mark=none,very thick,dashdotted,error bars/.cd,y dir=both,y explicit,x dir=both,x explicit,error bar style={solid,very thin}] table[x=time,x error= std-time, y=auc,y error=std-auc]  {Tikz/gene155_Sr.dat};
\legend{\algname\ (ours), \osr\ (ours),\ndsr\ (ours), \sr}
\end{axis}
\end{tikzpicture}     }
		\caption{Target CHEMBL100875}
		\label{fig:C2}
	\end{subfigure}
	
	\caption{Comparison of spectral regression methods and \algname\ for the 6 chosen targets of the ChEMBL data set. The AUC-ROC is given in function of the execution time in seconds. Converging to large AUC-ROC in small execution time (top left of the graph) is preferable.}
	\label{fig:chemblplots}
\end{figure*}
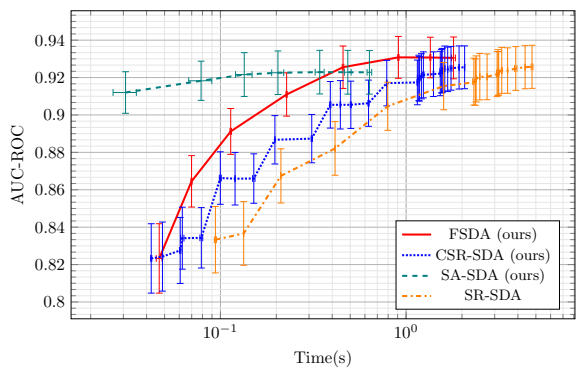
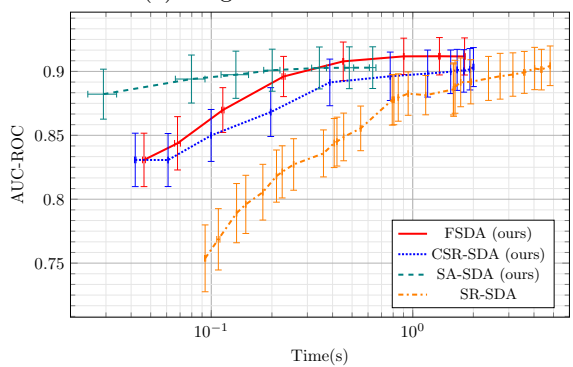
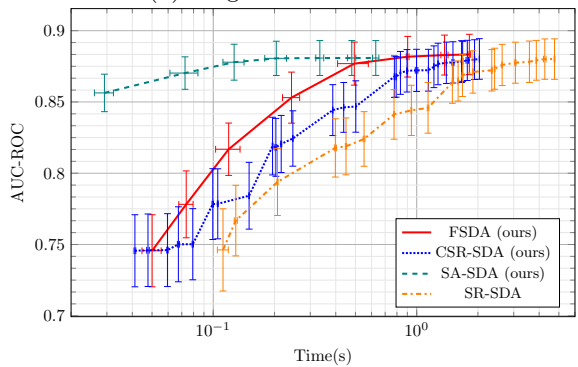
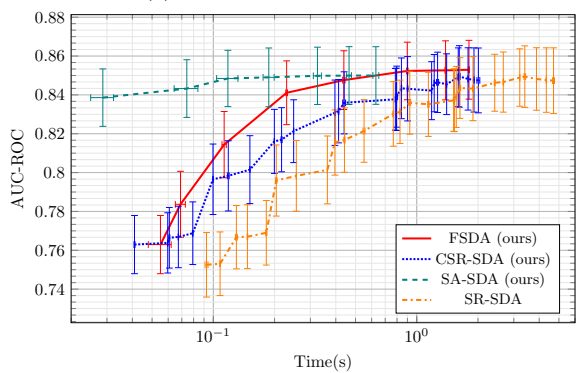
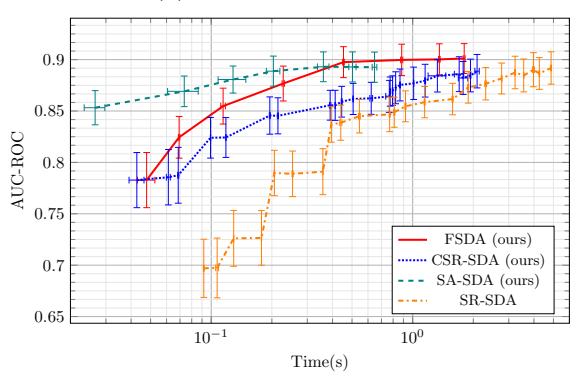

Figure \ref{fig:chemblplots} shows the AUC-ROC in function of the execution time for the different methods. Solving SDA directly in the feature space (\algname) gives higher or equal average AUC-ROC for all 6 targets in comparison with the spectral regression methods. There is no difference in AUC-ROC between \sr\ and \osr, but Figure \ref{fig:chemblplots} clearly shows that \osr\ requires less execution time to reach the same predictive performance. For this data, the AUC-ROC in function of execution times is higher for \algname\ than for \osr. Ideally, we prefer to reach high AUC-ROC with the smallest execution time (top left of the graph). We conclude that \algname\ is more suitable than \osr\ for the ChEMBL data set. Note that \ndsr\ reaches high AUC-ROC values in even less execution time since there is no regression step. Currently, we do not know if the difference in AUC-ROC after convergence between \osr, \ndsr\ and \algname\ is specifically for the data or inherently present due to the way the direction $\textbf{w}$ is calculated.  

\begin{table*}[ht]
	\renewcommand{\arraystretch}{1.0}
	\begin{center}
		\scalebox{1.0}{
			\begin{tabular}{c|c|c|c|c}\toprule
				$N_\beta$&$\beta$ &	CG & shifted CG& Speed-up\\ \midrule
				12&[$1\text{e-}9$, $1\text{e-}8$, $1\text{e-}7$, $1\text{e-}6$, $1\text{e-}5$, $1\text{e-}4$,  &43.11 s&6.02 s &7.16\\
				 &$1\text{e-}3$, $1\text{e-}2$, $1\text{e-}1$, $1$, $10$, $1\text{e}2$ ]& & &\\
				 12&[$1.0\text{e-}6$, $1.1\text{e-}6$, $1.2\text{e-}6$, $1.3\text{e-}6$, $1.4\text{e-}6$, $1.5\text{e-}6$,&50.98 s&5.68 s &8.98\\
				 &$1.6\text{e-}6$, $1.7\text{e-}6$, $1.8\text{e-}6$, $1.9\text{e-}6$, $2.0\text{e-}6$, $2.1\text{e-}6$]&  & &
\\ \bottomrule
		\end{tabular}}
	\end{center}
	\caption{The computation time of CG versus shifted CG for the regression phase using a tolerance of $10^{-3}$, the ChEMBL data matrix and given $\beta$-values.}\label{tab:shifted}
\end{table*}

Table \ref{tab:shifted} shows the execution time for CG and shifted CG solving the regularized least squares system in \osr\ (Algorithm \ref{alg:optimizedSR}, line 5) with a tolerance of $10^{-3}$ and several $\beta$-values. Using shifted CG leads to a speed-up of $7.16$ in computation time for $12$ $\beta$-values. Note that for the larger $\beta$-values the number of iterations of CG required to reach the desired tolerance is less than for the small $\beta$-values. Higher regularization parameters result in a smaller condition number and faster convergence for CG \cite{demmel1997applied}. Table \ref{tab:shifted} additionally shows the speed-up if $\beta$ would not vary significantly, resulting in a larger speed-up of $8.98$. If the $\beta$-values do not vary drastically, the number of iterations of CG do not change significantly and the achieved speed-up is larger.     
   
\subsection{Janssen data}\label{subsect:Janssen}
Next, we investigate the performance of the proposed methods on a industrial-scale data set containing a few million of compounds. This large-scale data set was provided by Janssen Pharmaceutica. Out of more than 1000 protein targets, we chose 6 with a large number of IC50 measurements, see Table \ref{tab:janssentargets} for an overview of these targets. For these targets, a compound was considered active with an IC50 value below 1.0 $\mu M$. Using ECFP over 10 million compound features were generated and the SDA similarity matrix was created using the Tanimoto similarity with a threshold of $0.4$.

\begin{figure*}
	\tikzsetnextfilename{Janssenplots2}
	\centering
	\resizebox{\linewidth}{!}{    \begin{tikzpicture}
        \begin{groupplot}[%
            group style={group name=janssen, group size= 3 by 2, horizontal sep=2.5cm, vertical sep=2.5cm},
            height=8cm,
            width=12cm,
            cycle list name=colors,
            %xmode = log,
            %ymode = log,
            %log y ticks with fixed point,
            grid=both,
            grid style={line width=.1pt, draw=gray!20},
            major grid style={line width=.2pt,draw=gray!60},
            minor tick num=5,
            xlabel=Time(s),ylabel=AUC-ROC,
            legend pos=south east,
            legend style={font=\Large},
           legend image code/.code={%
           	\draw[#1, draw=none, /tikz/.cd, bar width=3pt, yshift=-0.35em, bar shift=0pt] %
           	plot coordinates {(2*\pgfplotbarwidth,0.6em)};
           }, 
           legend style={
           	at={(0.0,-0.15)},
           	anchor=north west,
           	legend columns=-1,
           	/tikz/every even column/.append style={column sep=1.0cm}
           },
            y tick label style={
               	/pgf/number format/.cd,
               	fixed,
               	fixed zerofill,
               	precision=3,
               	/tikz/.cd
            },
        ]
        \nextgroupplot[%
        legend to name=zelda,
        mark=none,
        ]
        \coordinate (top) at (rel axis cs:0,1);% coordinate at top of the first plot
        \addplot+[mark=square,very thick,solid,error bars/.cd,y dir=both,y explicit,x dir=both,x explicit,error bar style={solid,very thin}] coordinates { 
        	(37.23388,0.9420582) +- (1.922870,0.001996436) };
        ;\addlegendentry{\algname\ (ours), 70 iterations}
        \addplot+[mark=*,fill=white,very thick,solid,error bars/.cd,y dir=both,y explicit,x dir=both,x explicit,error bar style={solid,very thin}] coordinates { 
        	(35.29530,0.9391852) +- ( 4.216344,0.001933802) };
        ;\addlegendentry{\osr\ (ours), 70 iterations}
        \addplot+[mark=pentagon,very thick,solid,error bars/.cd,y dir=both,y explicit,x dir=both,x explicit,error bar style={solid,very thin}] coordinates { 
        	(11.55797,0.9379854) +- ( 0.9748867,0.001344160) };
        ;\addlegendentry{\ndsr\ (ours), 70 iterations}
        \addplot+[mark=triangle,very thick,solid,error bars/.cd,y dir=both,y explicit,x dir=both,x explicit,error bar style={solid,very thin}] coordinates { 
        	(84.75218,0.9391475) +- ( 4.310016,0.001941351) };
        ;\addlegendentry{\sr, 70 iterations}
        
	  \nextgroupplot[mark=none]
	  \addplot+[mark=square,very thick,solid,error bars/.cd,y dir=both,y explicit,x dir=both,x explicit,error bar style={solid,very thin}]     coordinates { 
	  		(37.90586,0.9387998) +- (3.252597,0.004149372) };
	 \addplot+[mark=*,fill=white,very thick,solid,error bars/.cd,y dir=both,y explicit,x dir=both,x explicit,error bar style={solid,very thin}]     coordinates { 
	 		  	(36.15038,0.9350316) +- (4.514493,0.004353594) };
  \addplot+[mark=pentagon,very thick,solid,error bars/.cd,y dir=both,y explicit,x dir=both,x explicit,error bar style={solid,very thin}]     coordinates { 
 		  	  	(13.64853,0.9333941) +- (0.7126891,0.003927761) };
  \addplot+[mark=triangle,very thick,solid,error bars/.cd,y dir=both,y explicit,x dir=both,x explicit,error bar style={solid,very thin}]     coordinates { 
		  	  	  	(90.34999,0.9352667) +- (6.551342,0.004305642) };

	    \nextgroupplot[mark=none]
	    \addplot+[mark=square,very thick,solid,error bars/.cd,y dir=both,y explicit,x dir=both,x explicit,error bar style={solid,very thin}]     coordinates { 
	    	(39.39845,0.981643) +- (2.506893, 0.001393100) };
	    \addplot+[mark=*,fill=white,very thick,solid,error bars/.cd,y dir=both,y explicit,x dir=both,x explicit,error bar style={solid,very thin}]     coordinates { 
	    	(	42.03906,0.9787212) +- (5.507975,0.002277360) };
	    \addplot+[mark=pentagon,very thick,solid,error bars/.cd,y dir=both,y explicit,x dir=both,x explicit,error bar style={solid,very thin}]     coordinates { 
	    	(11.55941,0.9829297) +- (1.909335,0.001420246) };
	    \addplot+[mark=triangle,very thick,solid,error bars/.cd,y dir=both,y explicit,x dir=both,x explicit,error bar style={solid,very thin}]     coordinates { 
	    	(	71.01625,0.9780379) +- (11.17095,0.001684239) };

	    \nextgroupplot[mark=none]
	    \addplot+[mark=square,very thick,solid,error bars/.cd,y dir=both,y explicit,x dir=both,x explicit,error bar style={solid,very thin}]     coordinates { 
	    	(34.69718,0.8977146) +- (3.094835,0.01129341) };
	    \addplot+[mark=*,fill=white,very thick,solid,error bars/.cd,y dir=both,y explicit,x dir=both,x explicit,error bar style={solid,very thin}]     coordinates { 
	    	(	37.63691,0.8933242) +- (6.264090,0.01425923) };
	    \addplot+[mark=pentagon,very thick,solid,error bars/.cd,y dir=both,y explicit,x dir=both,x explicit,error bar style={solid,very thin}]     coordinates { 
	    	(12.20771,0.8855975) +- (1.045194,0.01299418) };
	    \addplot+[mark=triangle,very thick,solid,error bars/.cd,y dir=both,y explicit,x dir=both,x explicit,error bar style={solid,very thin}]     coordinates { 
	    	(89.4894,0.8933434) +- (9.449506,0.01441064) };
    	
	     \nextgroupplot[mark=none]
	     \addplot+[mark=square,very thick,solid,error bars/.cd,y dir=both,y explicit,x dir=both,x explicit,error bar style={solid,very thin}]     coordinates { 
	     	%(37.90586,0.9387998) +- (3.252597,0.004149372) };
	     	(37.37967,0.9874632) +- (4.456684,0.001270795) }; 
	     \addplot+[mark=*,fill=white,very thick,solid,error bars/.cd,y dir=both,y explicit,x dir=both,x explicit,error bar style={solid,very thin}]     coordinates { 
	     	(	43.36094,0.9854066) +- (2.21234,0.001689343) };
	     \addplot+[mark=pentagon,very thick,solid,error bars/.cd,y dir=both,y explicit,x dir=both,x explicit,error bar style={solid,very thin}]     coordinates { 
	     	(11.67433,0.9807554) +- (0.8357363,0.0005785129) };
	     \addplot+[mark=triangle,very thick,solid,error bars/.cd,y dir=both,y explicit,x dir=both,x explicit,error bar style={solid,very thin}]     coordinates { 
	     	(92.65434,0.9854818) +- (1.054402,0.001707471) };

        \nextgroupplot[mark=none]
        \addplot+[mark=square,very thick,solid,error bars/.cd,y dir=both,y explicit,x dir=both,x explicit,error bar style={solid,very thin}]     coordinates { 
        	(33.04031,0.9044805) +- (4.29118,0.009593832)};
        \addplot+[mark=*,fill=white,very thick,solid,error bars/.cd,y dir=both,y explicit,x dir=both,x explicit,error bar style={solid,very thin}]     coordinates { 
        	(	41.43653,0.8995109) +- (1.6696967,0.01053133) };
        \addplot+[mark=pentagon,very thick,solid,error bars/.cd,y dir=both,y explicit,x dir=both,x explicit,error bar style={solid,very thin}]     coordinates { 
        	(12.15362,0.8942240) +- (1.501033,0.01130920) };
        \addplot+[mark=triangle,very thick,solid,error bars/.cd,y dir=both,y explicit,x dir=both,x explicit,error bar style={solid,very thin}]     coordinates { 
        	(77.91339,0.8993248) +- (1.265756,0.01056883) };
        
        	%(39.42780,0.9609794) +- (0.9044797,0.009594134)};
        \coordinate (bot) at (rel axis cs:1,0);% coordinate at bottom of the last plot
        %;\addlegendentry{FSDA, Fixed iterations} % added space here for better alignment
        \end{groupplot} 
        \node[below = 1cm of janssen c1r1.south] {(a) Target 389};
        \node[below = 1cm of janssen c2r1.south] {(b) Target 685};
	    \node[below = 1cm of janssen c3r1.south] {(c) Target 739};
	    \node[below = 1cm of janssen c1r2.south] {(d) Target 1448};
        \node[below = 1cm of janssen c2r2.south] {(e) Target 1736};
        \node[below = 1cm of janssen c3r2.south] {(f) Target 1833};
         %\path (top)--(bot) coordinate[midway] (group center);
         \path (top|-current bounding box.north)--
         coordinate(legendpos)
         (bot|-current bounding box.north);
         \node[right=1em,inner sep=0pt] at([yshift=5ex,xshift=-14cm]legendpos) {\pgfplotslegendfromname{zelda}};  
         %\node[right=1em,inner sep=0pt] at(group center -| current bounding box.east) {\pgfplotslegendfromname{zelda}};     
         
    \end{tikzpicture}}
	\caption{Comparison of spectral regression methods and \algname\ for the 6 Janssen targets. The AUC-ROC is given in function of the execution time in seconds with a fixed number of iterations equal to $70$. Our methods improve the execution time by a factor $2$ in comparison with \sr\ or more for \ndsr. }
	\label{fig:Janssenplots2}
\end{figure*}
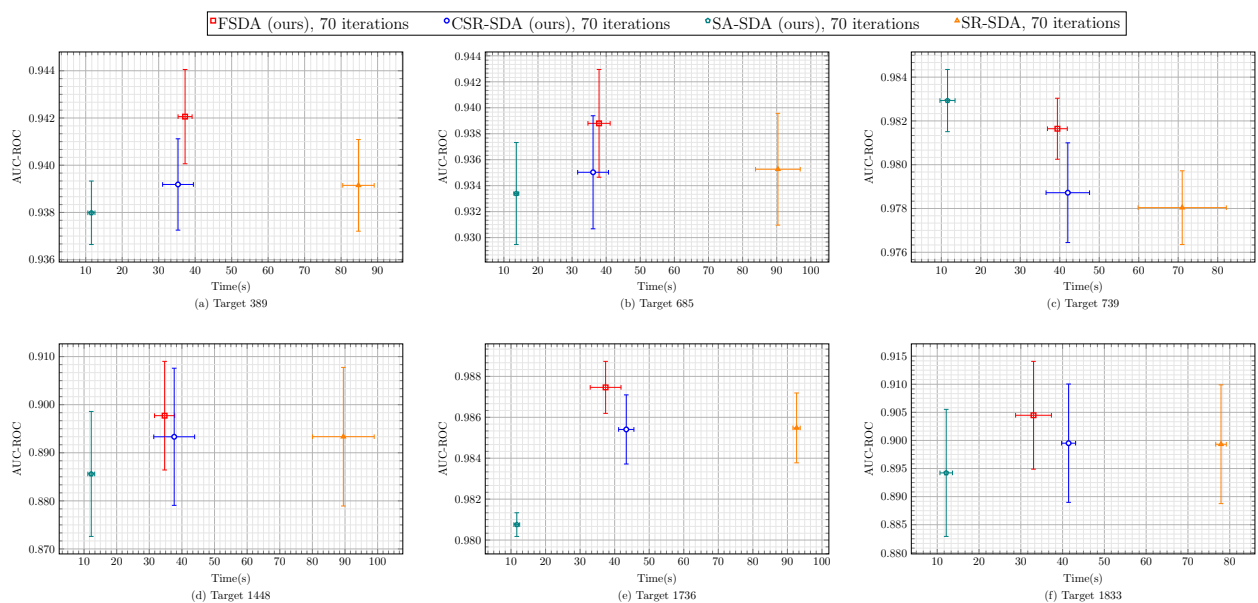

Figure \ref{fig:Janssenplots2} shows the AUC-ROC in function of the execution time for the targets. The number of iterations for the linear solves was set to $70$. As can be seen from the figure, the centralization improves the execution time by a factor 2 in comparison with \sr. \ndsr\ even gains a factor 7-8. All methods reach the same predictive performance with a small advantage for \algname\ as seen for the ChEMBL data set as well.  

\begin{table}[ht]
	\renewcommand{\arraystretch}{1.0}
	\begin{center}
		\scalebox{0.8}{
			\begin{tabular}{l|c|c|c|c}\toprule
				Name& Active& Inactive&Total labels&$\%$ labeled\\ \midrule
				389& 20000& 275000& 295000&15 \\			
				685& 10000&475000& 485000&25 \\ 		
				739& 5000&750000& 755000& 40\\ 									
				1448& 2500& 300000&302500 & 15\\ 								
				1736& 5000& 500000&505000 &25 \\ 				
				1833& 2500& 300000&302500 &15 \\ \bottomrule
			\end{tabular}
		}
	\end{center}
	\caption{The characteristics of the Janssen targets. }\label{tab:janssentargets}
\end{table}

\begin{figure*}
	\begin{subfigure}[b]{0.47\textwidth}
		\tikzsetnextfilename{JlB1}
		\centering
		\resizebox{\linewidth}{!}{\begin{tikzpicture}
\begin{axis}[
height=8cm,
width=12cm,
cycle list name=colors,
xmode = log,
grid=both,
grid style={line width=.1pt, draw=gray!20},
major grid style={line width=.2pt,draw=gray!60},
minor tick num=5,
%ymode = log,
%xmin=300,xmax=487615,ymin=0.5,ymax=1,
%xticklabels={,,},
%ytick={0.725,0.75,0.8,0.85,.9,0.950,0.9825},
y tick label style={
	/pgf/number format/.cd,
	fixed,
	fixed zerofill,
	precision=3,
	/tikz/.cd
},
xlabel=$\%$ Labeled,ylabel=AUC-ROC,
legend pos=south east
]
\addplot+[mark=none,very thick,solid,error bars/.cd,y dir=both,y explicit,error bar style={solid,very thin}] table[x=percent, y=auc,y error=std-auc] {Tikz/gene389_labels.dat};
\addplot+[mark=none,very thick,densely dashdotted,error bars/.cd,y dir=both,y explicit,error bar style={solid,very thin}] table[x=percent, y=auc01,y error=Std-auc01] {Tikz/gene389_labels.dat};
\addplot+[mark=none,very thick,densely dotted,error bars/.cd,y dir=both,y explicit,error bar style={solid,very thin}] table[x=percent, y=auc02,y error=Std-auc02] {Tikz/gene389_labels.dat};
\addplot+[mark=none,very thick,densely dashed,error bars/.cd,y dir=both,y explicit,error bar style={solid,very thin}] table[x=percent, y=auc03,y error=Std-auc03] {Tikz/gene389_labels.dat};
\addplot+[mark=none,very thick,loosely dashed,error bars/.cd,y dir=both,y explicit,error bar style={solid,very thin}] table[x=percent, y=auc04,y error=Std-auc04] {Tikz/gene389_labels.dat};
\addplot+[mark=none,very thick,loosely dashdotted,error bars/.cd,y dir=both,y explicit,error bar style={solid,very thin}] table[x=percent, y=auc05,y error=Std-auc05] {Tikz/gene389_labels.dat};
\legend{LDA,SDA$_{\alpha=0.1}$, SDA$_{\alpha=0.2}$, SDA$_{\alpha=0.3}$, SDA$_{\alpha=0.4}$, SDA$_{\alpha=0.5}$}
\end{axis}
\end{tikzpicture}   }
		\caption{Target 389}
		\label{fig:JlB1}
	\end{subfigure}
	\begin{subfigure}[b]{0.47\textwidth}
		\tikzsetnextfilename{JlA2}
		\centering
		\resizebox{\linewidth}{!}{\begin{tikzpicture}
\begin{axis}[
height=8cm,
width=12cm,
cycle list name=colors,
xmode = log,
grid=both,
grid style={line width=.1pt, draw=gray!20},
major grid style={line width=.2pt,draw=gray!60},
minor tick num=5,
%ymode = log,
%xmin=300,xmax=487615,ymin=0.5,ymax=1,
%xticklabels={,,},
%ytick={0.725,0.75,0.8,0.85,.9,0.950,0.9825},
y tick label style={
	/pgf/number format/.cd,
	fixed,
	fixed zerofill,
	precision=3,
	/tikz/.cd
},
xlabel=$\%$ Labeled,ylabel=AUC-ROC,
legend pos=south east
]
\addplot+[mark=none,very thick,solid,error bars/.cd,y dir=both,y explicit,error bar style={solid,very thin}] table[x=percent, y=auc,y error=std-auc] {Tikz/gene685_labels.dat};
\addplot+[mark=none,very thick,densely dashdotted,error bars/.cd,y dir=both,y explicit,error bar style={solid,very thin}] table[x=percent, y=auc01,y error=Std-auc01] {Tikz/gene685_labels.dat};
\addplot+[mark=none,very thick,densely dotted,error bars/.cd,y dir=both,y explicit,error bar style={solid,very thin}] table[x=percent, y=auc02,y error=Std-auc02] {Tikz/gene685_labels.dat};
\addplot+[mark=none,very thick,densely dashed,error bars/.cd,y dir=both,y explicit,error bar style={solid,very thin}] table[x=percent, y=auc03,y error=Std-auc03] {Tikz/gene685_labels.dat};
\addplot+[mark=none,very thick,loosely dashed,error bars/.cd,y dir=both,y explicit,error bar style={solid,very thin}] table[x=percent, y=auc04,y error=Std-auc04] {Tikz/gene685_labels.dat};
\addplot+[mark=none,very thick,loosely dashdotted,error bars/.cd,y dir=both,y explicit,error bar style={solid,very thin}] table[x=percent, y=auc05,y error=Std-auc05] {Tikz/gene685_labels.dat};
\legend{LDA,SDA$_{\alpha=0.1}$, SDA$_{\alpha=0.2}$, SDA$_{\alpha=0.3}$, SDA$_{\alpha=0.4}$, SDA$_{\alpha=0.5}$}
\end{axis}
\end{tikzpicture}   } 
		\caption{Target 685}
		\label{fig:JlA2}
	\end{subfigure}

	\begin{subfigure}[b]{0.47\textwidth}
		\tikzsetnextfilename{JlC1}
		\centering
		\resizebox{\linewidth}{!}{\begin{tikzpicture}
\begin{axis}[
height=8cm,
width=12cm,
cycle list name=colors,
xmode = log,
grid=both,
grid style={line width=.1pt, draw=gray!20},
major grid style={line width=.2pt,draw=gray!60},
minor tick num=5,
%ymode = log,
%xmin=300,xmax=487615,ymin=0.5,ymax=1,
%xticklabels={,,},
%ytick={0.725,0.75,0.8,0.85,.9,0.950,0.9825},
y tick label style={
	/pgf/number format/.cd,
	fixed,
	fixed zerofill,
	precision=3,
	/tikz/.cd
},
xlabel=$\%$ Labeled,ylabel=AUC-ROC,
legend pos=south east
]
\addplot+[mark=none,very thick,solid,error bars/.cd,y dir=both,y explicit,error bar style={solid,very thin}] table[x=percent, y=auc,y error=std-auc] {Tikz/gene739_labels.dat};
\addplot+[mark=none,very thick,densely dashdotted,error bars/.cd,y dir=both,y explicit,error bar style={solid,very thin}] table[x=percent, y=auc01,y error=Std-auc01] {Tikz/gene739_labels.dat};
\addplot+[mark=none,very thick,densely dotted,error bars/.cd,y dir=both,y explicit,error bar style={solid,very thin}] table[x=percent, y=auc02,y error=Std-auc02] {Tikz/gene739_labels.dat};
\addplot+[mark=none,very thick,densely dashed,error bars/.cd,y dir=both,y explicit,error bar style={solid,very thin}] table[x=percent, y=auc03,y error=Std-auc03] {Tikz/gene739_labels.dat};
\addplot+[mark=none,very thick,loosely dashed,error bars/.cd,y dir=both,y explicit,error bar style={solid,very thin}] table[x=percent, y=auc04,y error=Std-auc04] {Tikz/gene739_labels.dat};
\addplot+[mark=none,very thick,loosely dashdotted,error bars/.cd,y dir=both,y explicit,error bar style={solid,very thin}] table[x=percent, y=auc05,y error=Std-auc05] {Tikz/gene739_labels.dat};
\legend{LDA,SDA$_{\alpha=0.1}$, SDA$_{\alpha=0.2}$, SDA$_{\alpha=0.3}$, SDA$_{\alpha=0.4}$, SDA$_{\alpha=0.5}$}
\end{axis}
\end{tikzpicture}     }
		\caption{Target 739}
		\label{fig:JlC1}
	\end{subfigure}
	\begin{subfigure}[b]{0.47\textwidth}
		\tikzsetnextfilename{JlB2}
		\centering
		\resizebox{\linewidth}{!}{\begin{tikzpicture}
\begin{axis}[
height=8cm,
width=12cm,
cycle list name=colors,
xmode = log,
grid=both,
grid style={line width=.1pt, draw=gray!20},
major grid style={line width=.2pt,draw=gray!60},
minor tick num=5,
%ymode = log,
%xmin=300,xmax=487615,ymin=0.5,ymax=1,
%xticklabels={,,},
%ytick={0.725,0.75,0.8,0.85,.9,0.950,0.9825},
y tick label style={
	/pgf/number format/.cd,
	fixed,
	fixed zerofill,
	precision=3,
	/tikz/.cd
},
xlabel=$\%$ Labeled,ylabel=AUC-ROC,
legend pos=south east
]
\addplot+[mark=none,very thick,solid,error bars/.cd,y dir=both,y explicit,error bar style={solid,very thin}] table[x=percent, y=auc,y error=std-auc] {Tikz/gene1448_labels.dat};
\addplot+[mark=none,very thick,densely dashdotted,error bars/.cd,y dir=both,y explicit,error bar style={solid,very thin}] table[x=percent, y=auc01,y error=Std-auc01] {Tikz/gene1448_labels.dat};
\addplot+[mark=none,very thick,densely dotted,error bars/.cd,y dir=both,y explicit,error bar style={solid,very thin}] table[x=percent, y=auc02,y error=Std-auc02] {Tikz/gene1448_labels.dat};
\addplot+[mark=none,very thick,densely dashed,error bars/.cd,y dir=both,y explicit,error bar style={solid,very thin}] table[x=percent, y=auc03,y error=Std-auc03] {Tikz/gene1448_labels.dat};
\addplot+[mark=none,very thick,loosely dashed,error bars/.cd,y dir=both,y explicit,error bar style={solid,very thin}] table[x=percent, y=auc04,y error=Std-auc04] {Tikz/gene1448_labels.dat};
\addplot+[mark=none,very thick,loosely dashdotted,error bars/.cd,y dir=both,y explicit,error bar style={solid,very thin}] table[x=percent, y=auc05,y error=Std-auc05] {Tikz/gene1448_labels.dat};
\legend{LDA,SDA$_{\alpha=0.1}$, SDA$_{\alpha=0.2}$, SDA$_{\alpha=0.3}$, SDA$_{\alpha=0.4}$, SDA$_{\alpha=0.5}$}
\end{axis}
\end{tikzpicture}   } 
		\caption{Target 1448}
		\label{fig:JlB2}
	\end{subfigure}
	
	\begin{subfigure}[b]{0.47\textwidth}
		\tikzsetnextfilename{JlA1}
		\centering
		\resizebox{\linewidth}{!}{\begin{tikzpicture}
\begin{axis}[
height=8cm,
width=12cm,
cycle list name=colors,
xmode = log,
grid=both,
grid style={line width=.1pt, draw=gray!20},
major grid style={line width=.2pt,draw=gray!60},
minor tick num=5,
%ymode = log,
%xmin=300,xmax=487615,ymin=0.5,ymax=1,
%xticklabels={,,},
%ytick={0.725,0.75,0.8,0.85,.9,0.950,0.9825},
y tick label style={
	/pgf/number format/.cd,
	fixed,
	fixed zerofill,
	precision=3,
	/tikz/.cd
},
xlabel=$\%$ Labeled,ylabel=AUC-ROC,
legend pos=south east
]
\addplot+[mark=none,very thick,solid,error bars/.cd,y dir=both,y explicit,error bar style={solid,very thin}] table[x=percent, y=auc,y error=std-auc] {Tikz/gene1736_labels.dat};
\addplot+[mark=none,very thick,densely dashdotted,error bars/.cd,y dir=both,y explicit,error bar style={solid,very thin}] table[x=percent, y=auc01,y error=Std-auc01] {Tikz/gene1736_labels.dat};
\addplot+[mark=none,very thick,densely dotted,error bars/.cd,y dir=both,y explicit,error bar style={solid,very thin}] table[x=percent, y=auc02,y error=Std-auc02] {Tikz/gene1736_labels.dat};
\addplot+[mark=none,very thick,densely dashed,error bars/.cd,y dir=both,y explicit,error bar style={solid,very thin}] table[x=percent, y=auc03,y error=Std-auc03] {Tikz/gene1736_labels.dat};
\addplot+[mark=none,very thick,loosely dashed,error bars/.cd,y dir=both,y explicit,error bar style={solid,very thin}] table[x=percent, y=auc04,y error=Std-auc04] {Tikz/gene1736_labels.dat};
\addplot+[mark=none,very thick,loosely dashdotted,error bars/.cd,y dir=both,y explicit,error bar style={solid,very thin}] table[x=percent, y=auc05,y error=Std-auc05] {Tikz/gene1736_labels.dat};
\legend{LDA,SDA$_{\alpha=0.1}$, SDA$_{\alpha=0.2}$, SDA$_{\alpha=0.3}$, SDA$_{\alpha=0.4}$, SDA$_{\alpha=0.5}$}
\end{axis}
\end{tikzpicture}   }
		\caption{Target 1736}
		\label{fig:JlA1}
	\end{subfigure}
	\begin{subfigure}[b]{0.47\textwidth}
		\tikzsetnextfilename{JlC2}
		\centering
		\resizebox{\linewidth}{!}{\begin{tikzpicture}
\begin{axis}[
height=8cm,
width=12cm,
cycle list name=colors,
xmode = log,
grid=both,
grid style={line width=.1pt, draw=gray!20},
major grid style={line width=.2pt,draw=gray!60},
minor tick num=5,
%ymode = log,
%xmin=300,xmax=487615,ymin=0.5,ymax=1,
%xticklabels={,,},
%ytick={0.725,0.75,0.8,0.85,.9,0.950,0.9825},
y tick label style={
	/pgf/number format/.cd,
	fixed,
	fixed zerofill,
	precision=3,
	/tikz/.cd
},
xlabel=$\%$ Labeled,ylabel=AUC-ROC,
legend pos=south east
]
\addplot+[mark=none,very thick,solid,error bars/.cd,y dir=both,y explicit,error bar style={solid,very thin}] table[x=percent, y=auc,y error=std-auc] {Tikz/gene1833_labels.dat};
\addplot+[mark=none,very thick,densely dashdotted,error bars/.cd,y dir=both,y explicit,error bar style={solid,very thin}] table[x=percent, y=auc01,y error=Std-auc01] {Tikz/gene1833_labels.dat};
\addplot+[mark=none,very thick,densely dotted,error bars/.cd,y dir=both,y explicit,error bar style={solid,very thin}] table[x=percent, y=auc02,y error=Std-auc02] {Tikz/gene1833_labels.dat};
\addplot+[mark=none,very thick,densely dashed,error bars/.cd,y dir=both,y explicit,error bar style={solid,very thin}] table[x=percent, y=auc03,y error=Std-auc03] {Tikz/gene1833_labels.dat};
\addplot+[mark=none,very thick,loosely dashed,error bars/.cd,y dir=both,y explicit,error bar style={solid,very thin}] table[x=percent, y=auc04,y error=Std-auc04] {Tikz/gene1833_labels.dat};
\addplot+[mark=none,very thick,loosely dashdotted,error bars/.cd,y dir=both,y explicit,error bar style={solid,very thin}] table[x=percent, y=auc05,y error=Std-auc05] {Tikz/gene1833_labels.dat};
\legend{LDA,SDA$_{\alpha=0.1}$, SDA$_{\alpha=0.2}$, SDA$_{\alpha=0.3}$, SDA$_{\alpha=0.4}$, SDA$_{\alpha=0.5}$}
\end{axis}
\end{tikzpicture}   }
		\caption{Target 1833}
		\label{fig:JlC2}
	\end{subfigure}
	
	\caption{The AUC-ROC for the Janssen data in function of number of used labels (randomly subsampled) for 6 different targets. Using only $0.1\text{ to }1\%$ labels, SDA outperforms LDA for all the targets.}
	\label{fig:Janssenlabels}
\end{figure*}
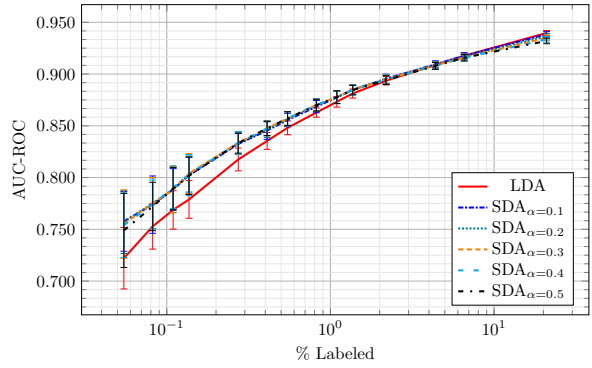
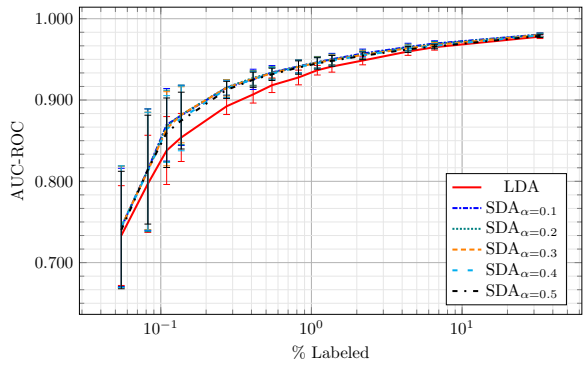
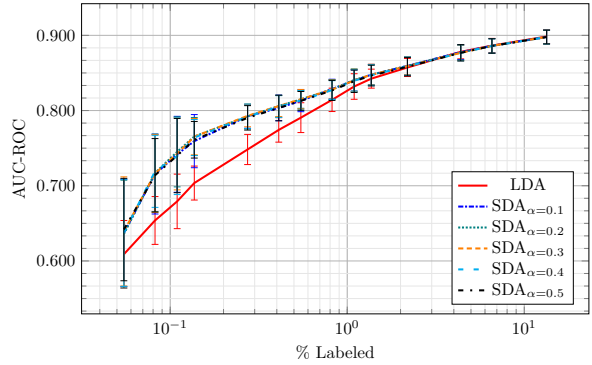
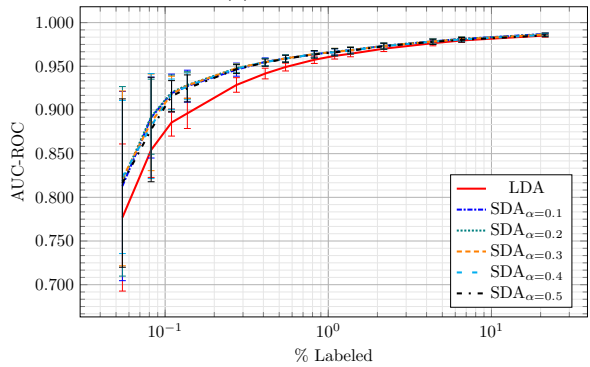
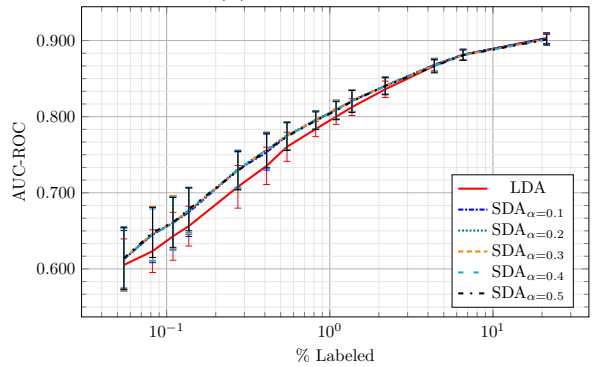

The relatively large assays for the Janssen data allows us to subsample the labeled data. Figure \ref{fig:Janssenlabels} gives the AUC-ROC in function of randomly sub-sampled labeled data for different values of the parameter $\alpha$. Using only $0.1\text{ to }1\%$ labels, SDA outperforms LDA for all the targets. Incorporating information of the unlabeled data trough SDA results in larger predictive performance if the number of labels is small.    

\section{Discussion}
Semi-supervised learning has mainly been developed for cases where there is a small number of labels. In our application, there is a medium number of labels and an abundance of unlabeled data samples. Further research is required specifically for this kind of large-scale data sets in the semi-supervised setting. The information of the unlabeled data is incorporated in the classification task. To include this information, SDA uses geometric distance in the feature space to define similarity between samples. Defining similarity in the feature space is sensitive to noise and redundant information of the defined features. 

For the chemogenomics application, the features are exact but do contain redundant information. The sample similarity could be further improved by removing non-informative features as suggested in \cite{wang2016semi}. Note that for our application, even the smallest speed-up for an algorithm is multiplied by the number of specific targets, resulting in a significant reduction of the total execution time. 
\section{Conclusion}
We have presented semi-supervised discriminant analysis for large-scale data. We have shown how to use the centralization to avoid the calculation of the non-discriminative eigenvector and an efficient way to implement the centralization. We used the shift-invariance property of Krylov subspaces to our advantage to solve SDA efficiently for different regularization parameters $\beta$. 

We have applied the spectral regression framework to SDA and showed how to centralize the data in spectral analysis. We additionally showed how to use SA as a standalone classification method in the semi-supervised setting. We applied SDA for target prediction in chemogenomics using data with medium number of labeled data and an abundance of unlabeled data. 

Using our proposed methods and implementations, it was possible to apply SDA on real-life large-scale data from the pharmaceutical industry in seconds, improving on previous state of the art by a factor 2 or more. Solving the problem directly in the feature space had experimentally for almost all targets better or equal predictive performance than the spectral regression methods.   
\section*{Conflict of interest}
No conflict of interest.
\section*{Acknowledgement}
The authors would like to thank the reviewers for their comments and suggestions. The authors would additionally like to thank Jean-Marc Neefs for annotating assays with protein and gene identifiers and Vladimir Chupakhin for helping with the compound normalization of the Janssen data. This work was supported by Flanders Innovation \& Entrepreneurship (IWT, No. IWT130406)  and Research Foundation - Flanders (FWO, No. G079016N). 
\section*{References}
\bibliographystyle{elsarticle-num}
\bibliography{references.bib} 
\end{document}